\documentclass[sn-mathphys-num]{sn-jnl}

\usepackage{graphicx}%
\usepackage{multirow}%
\usepackage{amsmath,amssymb,amsfonts}%
\usepackage{amsthm}%
\usepackage{mathrsfs}%
\usepackage[title]{appendix}%
\usepackage{xcolor}%
\usepackage{textcomp}%
\usepackage{manyfoot}%
\usepackage{booktabs}%
\usepackage{algorithm}%
\usepackage{algorithmicx}%
\usepackage{algpseudocode}%
\usepackage{listings}%

\usepackage{amssymb}
\usepackage{amsmath}
\usepackage[utf8]{inputenc}

\usepackage{microtype}

\usepackage{multirow}
\usepackage{multicol}
\usepackage{geometry}
\usepackage{inconsolata}
\usepackage{caption}
\usepackage{amsfonts}
\usepackage{booktabs}
\usepackage{xspace}
\usepackage{comment}
\usepackage{framed}
\usepackage{enumitem}
\usepackage{colortbl}

\usepackage{graphicx}
\usepackage{tcolorbox}
\usepackage{tcolorbox}

\theoremstyle{thmstyleone}%
\theoremstyle{thmstyletwo}%

\theoremstyle{thmstylethree}%

\begin{document}

\title[Article Title]{Evaluating Zero-Shot Multilingual Aspect-Based Sentiment Analysis with Large Language Models}







\author[1,2]{\fnm{Chengyan} \sur{Wu}}\email{chengyan.wu@m.scnu.edu.cn}
\equalcont{These authors contributed equally to this work.}
\author[3,4]{\fnm{Bolei} \sur{Ma}}\email{bolei.ma@lmu.de}
\equalcont{These authors contributed equally to this work.}
\author[5]{\fnm{Zheyu} \sur{Zhang}}\email{zheyu.zhang@tum.de}

\author[6]{\fnm{Ningyuan} \sur{Deng}}\email{dnylily@gmail.com}
\author*[6]{\fnm{Yanqing} \sur{He}}\email{heyq@istic.ac.cn}
\author*[1]{\fnm{Yun} \sur{Xue}}\email{xueyun@m.scnu.edu.cn}


\affil[1]{\orgdiv{Guangdong Provincial Key Laboratory of Quantum Engineering and Quantum Materials, School of Electronic Science and Engineering (School of Microelectronics)}, \orgname{South China Normal University}, \orgaddress{\city{Foshan}, \postcode{528225}, \country{China}}}

\affil[2]{\orgdiv{Guangdong Provincial Key Laboratory of Intelligent Information Processing},  \orgaddress{\city{Shenzhen}, \postcode{518060}, \country{China}}}

\affil[3]{\orgdiv{Department of Statistics}, \orgname{Ludwig Maximilian University of Munich}, \\ \orgaddress{ \city{Munich}, \postcode{80539}, \country{Germany}}}

\affil[4]{\orgdiv{Munich Center for Machine Learning},  \orgaddress{ \city{Munich}, \postcode{80539}, \country{Germany}}}

\affil[5]{\orgdiv{Technical University of Munich}, \orgaddress{\city{Munich}, \postcode{80333}, \country{Germany}}}

\affil[6]{\orgdiv{Research Center for Information Science Theory and Methodology}, \orgname{Institute of Scientific and Technical Information of China}, \orgaddress{\city{Beijing}, \postcode{100038}, \country{China}}}


\abstract{Aspect-based sentiment analysis (ABSA), a sequence labeling task, has attracted increasing attention in multilingual contexts. While previous research has focused largely on fine-tuning or training models specifically for ABSA, we evaluate large language models (LLMs) under zero-shot conditions to explore their potential to tackle this challenge with minimal task-specific adaptation. We conduct a comprehensive empirical evaluation of a series of LLMs on multilingual ABSA tasks, investigating various prompting strategies, including vanilla zero-shot, chain-of-thought (CoT), self-improvement, self-debate, and self-consistency, across nine different models. Results indicate that while LLMs show promise in handling multilingual ABSA, they generally fall short of fine-tuned, task-specific models. Notably, simpler zero-shot prompts often outperform more complex strategies, especially in high-resource languages like English. These findings underscore the need for further refinement of LLM-based approaches to effectively address ABSA task across diverse languages.}

\keywords{Multilingual Aspect-Based Sentiment Analysis, Large Language Model, Resource and Evaluation }



\maketitle

\section{Introduction}\label{sec1}

Aspect-based sentiment analysis (ABSA) is a sequence labeling task in natural language processing (NLP) focused on identifying fine-grained information in sentences, such as aspect terms and their associated sentiment polarity \cite{liu2012}. Although research \citep{ngcharacterising, zhu2021genotype, rahman2018processing} on ABSA has achieved success on English texts with large corpora, real-world social media interactions often involve multiple languages \citep{mao2022, zhang2021towards}, creating the need for multilingual sentiment analysis.

\begin{figure}[h]
    \centering
    \includegraphics[width=0.6\linewidth]{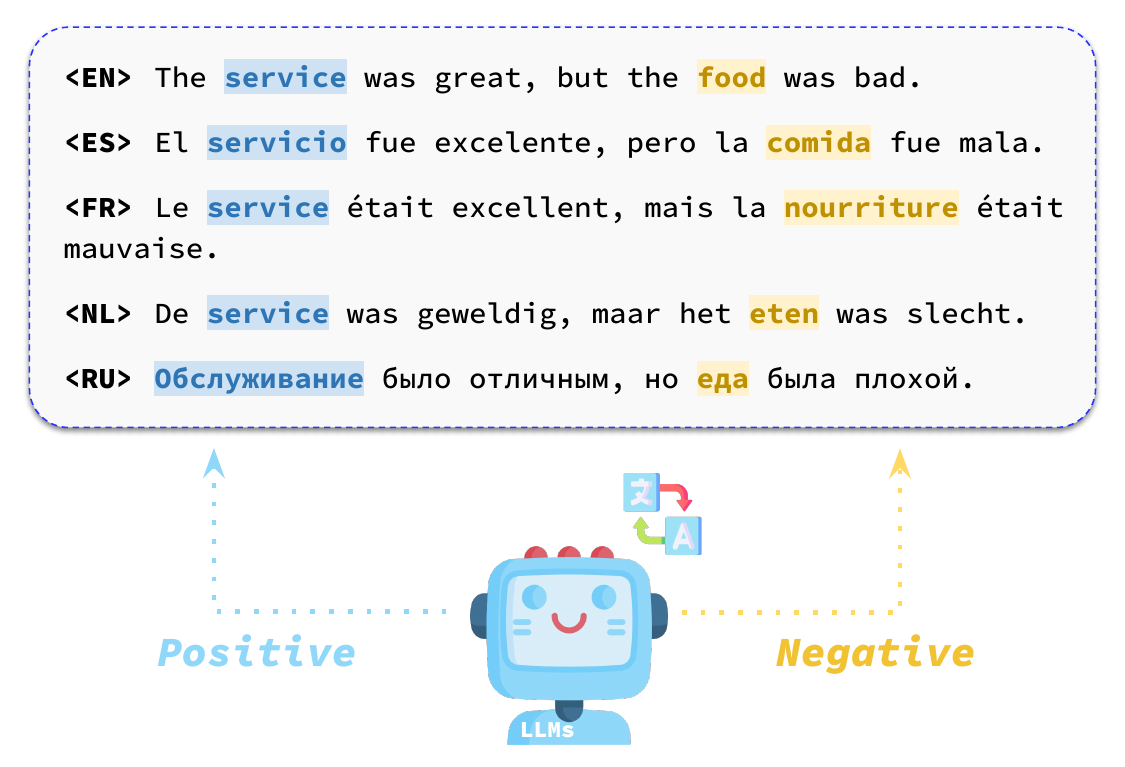}
    \caption{An example of multilingual ABSA task, where LLMs identify and analyze aspect terms across languages, determining the associated sentiments.}
    \label{fig:mABSA}
\end{figure}

Figure~\ref{fig:mABSA} illustrates a simple multilingual ABSA example, where the same restaurant review is presented in five different languages. A well-trained model should be able to identify the corresponding aspect terms \textit{service} and \textit{food} in each language and predict the associated sentiment polarity (\textit{service} is positive, \textit{food} is negative).

With the rise of LLMs, there has been growing interest in leveraging in-context learning approaches for various downstream tasks \citep[][\emph{inter alia}]{tran2024improving,van2024field,roumeliotis2024llms}.
Many studies have evaluated LLMs' capabilities in sequence labeling tasks such as named entity recognition (NER) \citep{xie-etal-2023-empirical} and part-of-speech (POS) tagging, particularly in multilingual settings \cite{ma-etal-2024-topro, nie2024decomposed}. Similarly, work on LLMs for sentiment analysis has included some assessment of ABSA task \cite{zhang-etal-2024-sentiment}. However, few studies have examined LLMs' performance on multilingual ABSA task specifically through various in-context learning strategies. As a result, the extent to which LLMs can effectively handle multilingual ABSA task under these conditions remains unclear. Understanding how well LLMs capture token-level aspect terms and their associated sentiments across languages will offer insights into their multilingual capabilities and aid in designing more effective models for specific tasks.

This work aims to comprehensively evaluate LLMs on multilingual ABSA tasks by exploring different prompt strategies and languages. Specifically, we aim to answer the following questions: (1) How do LLMs perform on multilingual ABSA task? (2) How do single-turn and multi-turn dialogue prompting paradigms affect the LLM performance on this task? (3) How do LLMs 
compare with smaller models trained on domain-specific datasets?

To address these questions, we begin by reviewing sequence labeling methods in the context of multilingual ABSA and large language models. Our evaluation spans datasets across multiple domains and five languages. While prior ABSA studies often use a single prompt strategy and language, this narrow evaluation fails to demonstrate LLMs' adaptability. Therefore, it is necessary to perform a more comprehensive evaluation using various prompt strategies and LLMs. 

Our experiments reveal several key findings: First, the LLM performance varies across languages, with higher performance in higher-resource languages. Second, when using single-turn and multi-turn dialogue methods requiring reasoning outputs, models show varying degrees of performance degradation, indicating the need for better solutions to optimize the reasoning process. Third, in extracting fine-grained structured information, LLMs still lag behind smaller language models trained on specific datasets. Finally, while LLM performance improves in few-shot settings with more demonstrations, it is constrained by context length, highlighting the need for more efficient solutions in the future.


\section{Related Work}\label{sec2}

\textbf{Multilingual ABSA.} 
Multilingual ABSA aims to identify aspects and their associated sentiment elements in the text across multiple languages. This task is complicated by linguistic diversity, cultural differences, and the scarcity of annotated datasets in non-English languages \citep{pontiki-etal-2016-semeval}. Early approaches relied heavily on supervised learning models, necessitating manually annotated data for each language \citep{lin2014cross, klinger2015instance, lambert2015aspect, barnes2016exploring}. This reliance on human resources to address data scarcity often made the annotation process extremely costly. Subsequent studies have addressed the limitation of insufficient low-resource sentiment analysis corpora by using machine translation methods. Recent advancements have shifted focus to cross-lingual embeddings and multilingual pre-trained language models, often fine-tuning these on specific languages for ABSA task \citep{9585242, xu2019bert, zhao2021knowledge}. \citet{zhang2021cross} improve the performance of cross-lingual sentiment analysis by introducing code-switched data and knowledge distillation. \citet{lin2023} propose a contrastive learning framework aimed at enhancing the semantic space representation in cross-lingual sentiment analysis tasks. While these approaches have enhanced generalization across languages, they still face difficulties when dealing with user-generated content, such as slang, abbreviations, and culturally dependent expressions \citep{zhang-etal-2021-cross}. The emergence of LLMs offers new opportunities to tackle these challenges, driving our investigation into their performance on such a nuanced multilingual task.

\textbf{LLMs for Sequence Labeling tasks.}
LLMs have recently shown promise in various sequence labeling tasks \citep{jun2023rating}, such as ABSA, NER \citep{zhang2023model}, and POS tagging \citep{xie-etal-2023-empirical, zhang-etal-2024-sentiment, wang2023gpt}. Unlike traditional models that often require task-specific architectures, LLMs benefit from extensive pre-training on diverse data, enabling them to provide flexible solutions across multiple tasks. Techniques such as prompting and fine-tuning have been explored to improve LLMs' applicability to sentiment analysis tasks \citep{simmering2023large, wang2023selfconsistency, wang-etal-2024-context}. \citet{Zhong_Ding_Liu_Du_Tao} find that the zero-shot performance of LLMs is comparable to that of fine-tuned BERT models. Meanwhile, \citet{Wang_Xie_Ding_Feng_Xia_2023} conduct a preliminary exploration on the application of ChatGPT to open-domain sentiment analysis and reasoning tasks. \citet{Han_Peng_Yang_Wang_Liu_Wan} conduct an extensive evaluation of ChatGPT on aspect-based sentiment analysis tasks, revealing a significant performance gap between ChatGPT and other state-of-the-art methods. \citet{Deng_Bashlovkina_Han_Baumgartner_Bendersky} study the use of pseudo-label data generated by LLMs to fine-tune smaller models, achieving the performance of supervised learning models. However, these efforts have generally been limited to monolingual (English) sentiment analysis tasks, often involving datasets from different domains, which causes the performance of models to fluctuate due to domain-specific differences, complicating the interpretation of these variations. Besides, several challenges remain, particularly when adapting LLMs to multilingual contexts and balancing trade-offs between model size, computational cost, and performance \citep{simmering2023large}. Exploring LLMs on multilingual parallel corpora can help mitigate data-level differences and allow for a more focused evaluation of model performance across different languages. In our work, we mainly extend this research by evaluating LLMs' effectiveness on multilingual ABSA, a complex task that requires handling both aspect extraction and sentiment classification across languages.

\section{Prompting Strategies}
\label{sec:prompting}

\begin{figure}
    \centering
    \includegraphics[width=1\linewidth]{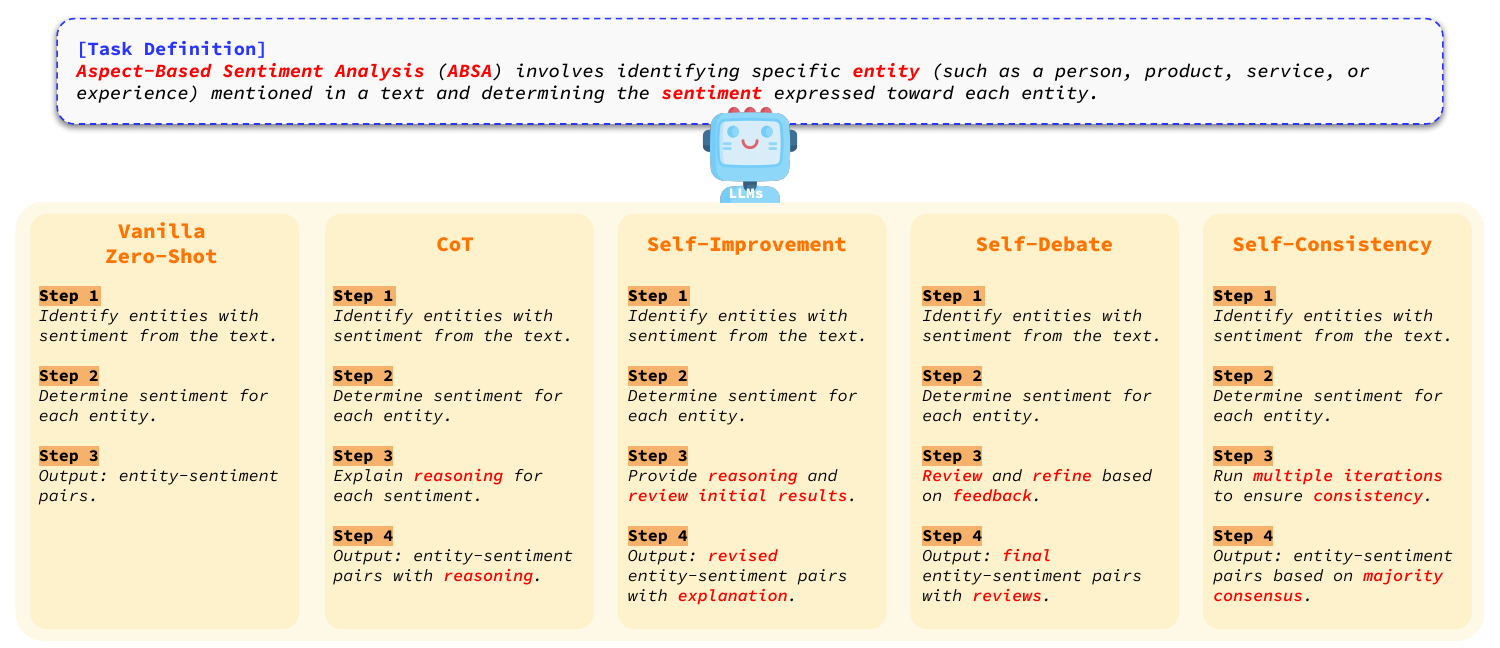}
    \caption{Prompting Strategies used for the experiments.}
    \label{fig:prompts}
\end{figure}

To evaluate the ability of LLMs to perform multilingual ABSA task, we employ a range of prompting strategies, from basic zero-shot prompts to more complex multi-turn reasoning prompts. Given that LLMs are highly sensitive to prompts and may generate varying responses even when the prompts are semantically similar \cite{perez2021, lu-etal-2022-fantastically, zhang-etal-2024-sentiment}, we ensure consistency across each prompting strategy. Figure \ref{fig:prompts} illustrates five prompting strategies assessed in this study. In all cases, we include a short introduction sentence in the beginning to make the model familiar with the task. Considering the token-level nature of the ABSA task, we include formatting instructions and explicitly instruct the models to output the results in a JSON format, containing the aspect term (entity) and its associated sentiment, in the prompts. The expected output structure is as follows:

{\small
\begin{verbatim} {"entity": "<entity>", "sentiment": "<label>"} \end{verbatim}
}

\textbf{Vanilla Zero-Shot.} The vanilla zero-shot prompt directly asks for the aspect terms and their sentiments, with a given input sentence. This is the major prompting method conducted in \cite{zhang-etal-2024-sentiment}. 

\textbf{CoT.} Chain-of-Thought (CoT) prompting encourages the model to generate a sequence of intermediate reasoning steps before arriving at a final answer, which can significantly enhance its ability to handle complex, multi-step reasoning tasks \cite{wei2022, wang-etal-2023-towards}. In this study, CoT prompting extends the vanilla zero-shot approach by explicitly instructing the model to articulate its reasoning process, thereby providing greater transparency into the decision-making pathway.

\textbf{Self-Improvement.} Drawing from the self-improving framework, which uses unlabeled data and self-annotation and retrieving to enhance performance in zero-shot tasks such as NER \cite{xie-etal-2024-self}, we introduce a simplified self-improvement prompting strategy.  The self-improvement prompting strategy is a three-step, multi-turn approach designed to refine the model's initial output. The first step involves standard zero-shot prompting to generate an initial response. In the second step, the model is asked to explain its reasoning behind the initial response, encouraging introspection and evaluation of its own outputs. Finally, the model is prompted with both its initial response and its reasoning to generate a revised, more informed answer. This iterative process helps the model to self-correct and produce revised results.

\textbf{Self-Debate.} The self-debate prompting strategy introduces a simulated dialogue \citep{wang-etal-2024-rethinking-bounds,zhang-etal-2024-exploring} between multiple agents within the model. It begins with a zero-shot prompt to generate an initial response, as in the self-improvement approach. In the second step, the model is prompted to assume the role of a second agent tasked with critically reviewing and, if necessary, correcting the initial response. In the third step, the model acts as a third agent that evaluates both preceding outputs and engages in a simulated debate to arrive at a final decision. This approach can be iterated over multiple rounds to further explore the model's capability for self-assessment and refinement, although for efficiency, we limit this process to three steps.

\textbf{Self-Consistency.} The self-consistency prompting strategy aims to stabilize the model's outputs by running the same experiment multiple times ($n$ iterations) with varying temperature settings. This approach leverages sampling diversity to generate multiple potential outputs, from which the most frequently occurring result is selected as the final answer, which is called majority voting. The empirical evaluation shows that self-consistency boosts the performance of CoT prompting on a few reasoning benchmarks \cite{wang2023selfconsistency}. This self-consistency is also tested to have improved performance on zero-shot NER tasks \cite{xie-etal-2023-empirical}.

\section{Experimental Setup}

\textbf{Models.} To conduct a comprehensive evaluation, we select a diverse range of open-source and closed-source instruction-tuned LLMs to perform the ABSA task: Llama-3.1 8B \cite{llama31modelcard}, Mistral 7B \cite{jiang2023mistral7b}, Gemma-2 9B \cite{gemmateam2024gemma2improvingopen}, Qwen-2.5 7B \cite{qwen2.5}, Zephyr 7B \cite{tunstall2023zephyrdirectdistillationlm}, 
Phi-3.5-mini 3.8B \cite{abdin2024phi3technicalreporthighly}, Gemini-1.5 \cite{geminiteam2024gemini15unlockingmultimodal}, Claude-3.5 \cite{anthropic2024claude35}, and GPT-4o \cite{gpt40modelcard}. These models are chosen for their diverse architectures and tuning strategies, offering a broad assessment of both state-of-the-art open-source and proprietary LLM capabilities.

\textbf{Decoding Temperature.} For all LLMs used in our experiments, unless stated otherwise, we set the decoding temperature to 0.0, employing a greedy decoding strategy to minimize random variations in the results. In the Self-Consistency setup, where the goal is to identify the most consistent output, we experiment with a range of temperatures \{0.2, 0.4, 0.6, 0.8, 1.0\}, in addition to the default 0.0 setting.

\textbf{Dataset.} We use the SemEval-2016 dataset \citep{pontiki-etal-2016-semeval} to perform a comprehensive evaluation of the selected models. This dataset consists of real user reviews across eight languages, with ABSA annotations available for English (EN), French (FR), Spanish (ES), Dutch (NL), Russian (RU), and Turkish (TK). However, due to the limited size of the Turkish test set (fewer than 150 sentences), we exclude it from our evaluation, consistent with prior multilingual ABSA studies \cite{zhang-etal-2021-cross, lin2023}. The data statistics are shown in Appendix \ref{sec:dataset}. 

\textbf{Baselines.} Due to the limited research on multilingual or cross-lingual ABSA task in the past two years, we adopt commonly used supervised fine-tuning baselines, as used in \cite{zhang-etal-2021-cross, lin2023}. Specifically, we fine-tune two multilingual models, mBERT \cite{devlin-etal-2019-bert} and XLM-R \cite{conneau-etal-2020-unsupervised}, on the multilingual data set as baselines. These models provide strong points of comparison for our zero-shot and few-shot LLM prompting strategies. Detailed parameter settings for the fine-tuning process are provided in Appendix \ref{sec:ft_parameter}.

\textbf{Evaluation Metric.} We evaluate five languages using the Micro-F1 metric on LLMs, which is well-suited for tasks requiring both aspect and sentiment prediction. In ABSA task, a prediction is considered correct only if both the aspect term and its associated sentiment match the ground truth. This ensures that the evaluation reflects the model’s ability to accurately identify both components.

\section{Results}

\subsection{Overall Observation}
\label{sec:overall}
\begin{table*}[ht]
\setlength\tabcolsep{5pt}
\centering
\tiny
\begin{tabular}{ll|ccccccccc|c}
\toprule 
Methods&Lang.& Claude-3.5 & Gemma-2 & Gemini-1.5 & GPT-4o & Llama-3.1 & Mistral &  Phi-3.5 & Qwen-2.5 & Zephyr& Avg.\\
\midrule
\multirow{6}{*}{Zero-Shot}
& EN    & 53.37  & 53.95  & 60.39 & 55.81  & 45.60 & 48.26   & 52.19& 54.64& 39.15  & 51.49\\
& ES    & 44.36  & 48.29  & 48.80 & 49.91  & 32.49 & 38.32   & 40.08& 48.10& 33.48  & 42.65\\
& FR    & 42.73  & 48.88  & 50.94 & 48.43  & 23.15 & 37.21   & 39.81& 43.60& 32.21  & 40.77\\
& NL    & 42.03  & 46.75  & 50.24 & 49.94  & 33.53 & 33.98   & 37.41& 42.59& 26.79  & 40.36\\
& RU    & 33.90  & 40.25  & 39.34 & 45.15  & 30.18 & 26.58   & 28.71& 37.15& 22.67  & 33.77\\
\cmidrule(l){2-12}
& Avg.  & 43.28  & 47.63  & 49.94 & 49.85  & 32.99 & 36.87   & 39.64& 45.22& 30.86  & 41.81\\
\midrule
\multirow{6}{*}{CoT} 
& EN    & 45.14  & 56.08  & 55.85 & 59.84  & 45.95 & 44.14   & 47.81& 52.62& 44.50  & 50.22\\
& ES    & 40.58  & 48.99  & 45.38 & 54.01  & 32.63 & 33.85   & 37.58& 44.21& 32.81  & 41.12\\
& FR    & 34.50  & 49.80  & 43.27 & 52.28  & 23.77 & 32.10   & 37.40& 38.36& 33.14  & 38.29\\
& NL    & 37.84  & 47.73  & 46.78 & 53.06  & 34.83 & 27.43   & 31.18& 39.36& 29.28  & 38.61\\
& RU    & 34.31  & 40.56  & 38.14 & 44.88  & 31.17 & 25.86   & 24.79& 31.87& 26.24  & 33.09\\
\cmidrule(l){2-12}
& Avg.  & 38.48  & 48.63  & 45.88 & 52.81  & 33.67 & 32.68   & 35.75& 41.28& 33.19  & 40.26\\
\midrule 
\multirow{6}{*}{Self-Improve.} 
& EN    & 51.69  & 47.69  & 53.07 & 50.18  & 38.83 & 38.10   & 48.68& 49.14& 40.61  & 46.44\\
& ES    & 42.69  & 42.56  & 43.66 & 42.83  & 31.10 & 31.50   & 38.29& 40.84& 34.38  & 38.65\\
& FR    & 42.05  & 42.42  & 45.70 & 41.09  & 21.72 & 29.80   & 37.87& 38.23& 32.75  & 36.85\\
& NL    & 38.52  & 37.53  & 42.91 & 37.17  & 30.71 & 26.20   & 34.68& 34.72& 27.45  & 34.43\\
& RU    & 34.26  & 33.51  & 35.05 & 34.69  & 26.86 & 21.23   & 27.03& 29.11& 23.73  & 29.50\\
\cmidrule(l){2-12}
& Avg.  & 41.84  & 40.74  & 44.08 & 41.19  & 29.84 & 29.37   & 37.31& 38.41& 31.78  & 37.17\\
\midrule 
\multirow{6}{*}{Self-Debate} 
& EN    & 51.44  & 40.93  & 46.86 & 45.86  & 25.30 & 27.00   & 45.88& 42.59& 35.78  & 40.18\\
& ES    & 40.03  & 36.89  & 37.70 & 39.45  & 18.67 & 22.69   & 37.08& 37.05& 31.33  & 33.43\\
& FR    & 39.02  & 36.90  & 40.94 & 38.11  & 15.43 & 23.80   & 34.54& 32.02& 27.36  & 32.01\\
& NL    & 38.41  & 34.79  & 37.67 & 36.24  & 20.79 & 19.40   & 32.14& 29.81& 23.36  & 30.29\\
& RU    & 31.58  & 29.95  & 32.39 & 32.60  & 19.82 & 16.62   & 24.33& 25.43& 19.86  & 25.84\\
\cmidrule(l){2-12}
& Avg.  & 40.10  & 35.89  & 39.11 & 38.45  & 20.00 & 21.90   & 34.79& 33.38& 27.54  & 32.35\\
\midrule 
\multirow{6}{*}{Self-Consist.} 
& EN    & 48.21  & 49.95  & 56.07 & 56.60  & 38.18 & 42.54   & 43.18& 52.09& 29.29  & 46.23\\
& ES    & 36.84  & 44.49  & 41.25 & 46.28  & 24.93 & 29.04   & 29.08& 44.58& 22.06  & 35.39\\
& FR    & 38.84  & 45.98  & 44.13 & 47.25  & 21.87 & 27.55   & 30.07& 41.82& 20.09  & 35.29\\
& NL    & 34.72  & 43.31  & 42.68 & 45.61  & 25.10 & 24.46   & 25.88& 39.66& 16.25  & 33.07\\
& RU    & 29.97  & 36.56  & 34.73 & 41.62  & 24.18 & 19.74   & 20.85& 33.97& 14.67  & 28.48\\
\cmidrule(l){2-12}
& Avg.  & 37.72  & 44.06  & 43.77 & 47.47  & 26.85 & 28.67   & 29.81& 42.42& 20.47  & 35.69 \\
\bottomrule 
\end{tabular}
\caption{Performance comparison of various methods on different languages.}
\label{tab:performance}
\end{table*}

Table \ref{tab:performance} summarizes the performance of five methods, including zero-shot, CoT, self-improvement, self-debate, and self-consistency methods on the multilingual ABSA task. For each method, we report the performance across five different languages, as well as the average performance, to provide a clearer comparison of the models' effectiveness under different approaches. For multi-turn dialogues, we select the output from the last turn as the final result. 

\textbf{Prompt Sensitivity Across Models, Methods, Languages in Performance.} 
Different prompt designs can lead to significant performance variations. To fairly investigate the impact of this sensitivity on the multilingual ABSA task, we establish uniform performance variations for prompt evaluation for each method. 
For fine-grained structured outputs, the impact of prompts on performance under the same prompt is significant. Generally, we observe that simple zero-shot prompts lead to better performance than other complex prompts in most cases. 
In terms of overall performance, GPT-4o achieves the best average performance in all five methods, reaching an F1 score of 45 95\%, which can be attributed to the stronger capabilities of the model and its use of a more diverse multilingual corpus during pre-training, leading to better handling of downstream tasks. When comparing the performance between the languages, we clearly observe that the results on English are the best, followed by Spanish and French, and the results on Russian are generally the worst. This is not surprising, as current LLMs are mostly English-centric \cite{nie2024decomposed} and we apply English prompts during the experiments. Additionally, the relatively higher performance on Spanish and French can be partially attributed to their linguistic similarity with English, particularly in terms of vocabulary, syntax, and shared Latin roots, which may facilitate transfer learning in multilingual settings. In contrast, Russian is typologically more distant from English, with distinct morphological and syntactic structures, which likely contributes to the lower performance. Moreover, tokenization and subword segmentation may be less optimal for languages with rich morphology like Russian, further degrading model performance. These observations highlight that cross-lingual transfer is not uniform and that language-specific properties play a crucial role in model effectiveness. Details about the prompts can be found in the Appendix \ref{sec:Prompts}.

\textbf{Challenges 
in CoT Reasoning for Multilingual ABSA.} 
For single-turn dialogues, we can observe that the zero-shot method significantly outperforms the CoT method across all benchmarks (with an average 1.55\% Micro-F1 difference). In multilingual ABSA task, the quality and structure of the dataset are critical to the performance of sentiment classification models, especially when we apply step-by-step reasoning methods. However, previous studies \citep{WANG2024106315} have shown that the SemEval-2016 dataset \citep{pontiki-etal-2016-semeval} contains a significant number of sentences with incomplete structures, which lack explicit sentiment-bearing entities or grammatical completeness, making it difficult to find explicit evidence from the sentence. When encountering incomplete sentences, the CoT method may attempt to "force" the generation of reasoning steps, leading to over-extrapolation or reasoning errors. 

\begin{figure}[htbp]
\centering
  \includegraphics[width=0.6\columnwidth]{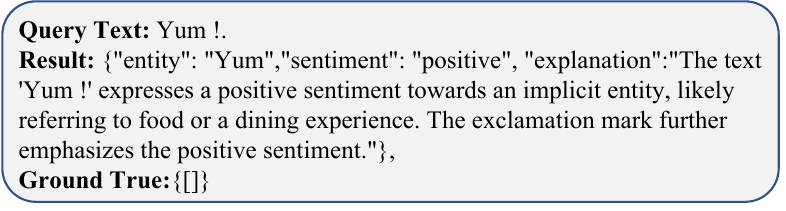}
  \caption{Case study of an incomplete sentence in CoT scenario. Results are extracted from the Claude-3.5 output.}
  \label{fig:cot-zeroshot compare}
\end{figure}

\begin{figure}[htbp]
\centering
  \includegraphics[width=0.6\columnwidth]{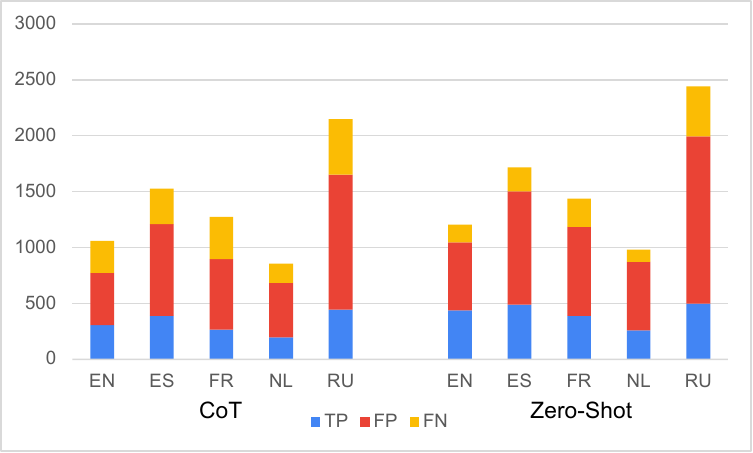}
  \caption{Comparison of True Positives, False Positives, and False Negatives of Claude-3.5 outputs in Zero-Shot and CoT Methods.}
  \label{fig:cot&zeroshot}
\end{figure}

\begin{table}[htbp]
\centering
\begin{tabular}{ccccccc}
\toprule
& EN & ES & FR & NL & RU & Avg. \\
\midrule
mBERT  &66.68& 67.97 &62.78  & 56.31 & 59.49 & 62.65 \\
XLM-R &72.92& 72.74 &67.26 &  64.72 & 66.66 & 68.86 \\
\midrule
GPT-4o-CoT  &59.84& 54.01 & 52.28 & 53.06 & 44.88 & 52.81 \\
\bottomrule
\end{tabular}
\caption{The fully-supervised baseline results, compared with GPT-4o-CoT results.}
\label{tab:baseline}
\end{table}
For example, we present a representative example in Figure \ref{fig:cot-zeroshot compare} where the sentence is composed of phrases and punctuation marks with a positive sentiment polarity. The CoT method fails to identify explicit evidence within the sentence, leading to over-reasoning and ultimately resulting in an incorrect inference. Moreover, CoT methods not only require the model to identify sentiment-polarized entities but also to generate a detailed reasoning process. However, the model often produces incorrect reasoning chains or fails to find reasoning evidence when the sentences lack clear sentiment clues, leading to entity omission. In Figure \ref{fig:cot&zeroshot}, we present the statistics of true positives, false positives, and false negatives for Claude-3.5 under both zero-shot and CoT methods, further illustrating our conclusions. Compared to the zero-shot method, the CoT method exhibits a significant decrease in the number of correct predictions and over-predictions, as well as a noticeable increase in the number of entity omissions, which aligns with our hypothesis.

\vspace{5pt}
\textbf{Multi-Turn vs. Single-Turn in Performance.}
As seen in Table \ref{tab:performance}, compared to the zero-shot single-turn dialogue method, the multi-turn dialogue method (self-improvement \& self-debate) exhibits a significant performance drop across all languages. The best average performance achieves an F1 score of 37.17\%, which is 4.64\% lower than that of the zero-shot method. The performance variation indicates that while multi-turn dialogues can enrich the interaction by allowing for more detailed exchanges, they may inadvertently lead to misunderstandings or misinterpretations of the sentiment expressed. This is particularly evident in languages where sentiment cues are heavily context-dependent. We will delve deeper into these findings and discuss the potential implications of adopting multi-turn dialogue strategies for multilingual aspect-based sentiment analysis in \S\ref{sec:multi-diog}. 

\textbf{Self-Consistency in Different Temperatures.}
Our analysis of self-consistency in multilingual ABSA task reveals a decrease in performance when averaging the outputs from five different temperature settings (0 to 1) compared to using a singular temperature of 0. The average performance decreased by 5.26\%, 7.26\%, 5.48\%, 7.29\%, and 5.29\% respectively in five languages compared to the initial zero-shot setting (greedy decoding, temperature=0), indicating that while the varying temperature can introduce diversity in responses, it may also lead to inconsistencies that negatively impact overall performance. These results emphasize the importance of carefully selecting temperature settings to maintain output stability in multilingual contexts. 

\subsection{Comparison with Fine-Tuning Baseline}
\label{sec:baselines}
The comparison between prompt-engineering methods with LLMs and fine-tuning method reveals distinct performance trends in ABSA task. As shown in Table \ref{tab:performance}, we can observe that the prompt-engineering method with LLMs struggles to extract fine-grained structured aspect terms and corresponding sentiment compared to fine-tuning methods. 
We show the fully-supervised baseline results from the mBERT and XLM-R models in Table \ref{tab:baseline}, as well as the results from the best-performing prompting method and the model (GPT-4o-CoT). 
We notice, even the best-performing LLM prompting method (GPT-4o-CoT) only reaches around 77\% of the average performance achieved by fine-tuned models (compared to XLM-R). This performance gap becomes especially obvious in languages with complex morphological structures and domain-specific datasets such as Russian and French, where fine-tuned models, leveraging task-specific training, exhibit more consistent and reliable results. The uniqueness of prompting lies in its flexibility and adaptability, but applying it effectively in specific scenarios can be challenging. This is because prompt-engineering approaches often require carefully crafted instructions that can vary significantly depending on the task, making consistent performance harder to achieve across diverse settings.

\subsection{The Effect of Model Size in Performance}
\label{sec:model-size}

To explore how model size affects performance in ABSA tasks, we conduct experiments using different sizes of the Qwen series models (Qwen2.5 with 0.5B, 1.5B, 7B, 14B, 32B, 72B). In our experimental setup, we focus on representative zero-shot and self-improvement methods in both single-turn and multi-turn dialogues. As shown in Table \ref{tab:result_model_size}, we observe the following findings: 1) Model performance increases with size. For example, the Qwen2.5 14B model outperforms the Qwen2.5 1.5B model by 21.98\% and 22.23\% in average F1 score for zero-shot and self-improvement methods, respectively. 2) Zero-Shot performs better overall than Self-Improvement, which aligns with the trends reported in Table \ref{tab:performance}. In conclusion, larger language models, having been exposed to more data and more complex training frameworks during pretraining, achieve better performance in multilingual ABSA tasks. This also suggests that, for specific downstream tasks, selecting a larger model with superior performance as the base model can be beneficial.
\begin{table}[htbp]
\centering
\begin{tabular}{c|c|cccccc}
\toprule
Methods&Model & EN & ES & FR & NL & RU & Avg. \\
\midrule
&Qwen-2.5 0.5B  &10.24& 8.97 & 9.88 & 6.82 & 4.06 & 7.99 \\
&Qwen-2.5 1.5B  &41.26& 32.75 &25.87  & 23.01 & 15.48 & 27.67 \\
\multicolumn{1}{c|}{Zero-Shot}&Qwen-2.5 7B &54.64& 48.10 &43.60 &  42.59 & 37.15 & 45.22 \\
&Qwen-2.5 14B  &60.82& 52.41 & 47.77 & 46.53 & 40.72 & 49.65 \\
&Qwen-2.5 32B &64.75&55.62&55.46&57.17&42.77&55.50 \\
&Qwen-2.5 72B &67.89&59.77&57.36&60.33&46.69&58.41 \\
\midrule
&Qwen-2.5 0.5B  &7.88& 5.49 & 6.98 & 4.68 & 3.03 & 5.61 \\
&Qwen-2.5 1.5B  &33.43& 27.65 &21.52  & 17.91 & 12.33 & 22.57 \\
\multicolumn{1}{c|}{Self-Improve.}&Qwen-2.5 7B &49.14& 40.84 &38.23 &  34.72 & 29.11 & 38.41 \\
&Qwen-2.5 14B  &56.47& 45.78 & 42.94 & 43.13 & 35.68 & 44.80 \\
&Qwen-2.5 32B &58.39&52.25&50.79&52.08&38.82&50.47\\
&Qwen-2.5 72B &61.35&55.67&54.84&55.37&41.28&53.70\\
\bottomrule
\end{tabular}
\caption{The results compared with different sizes of models under zero-shot and self-improvement settings.}
\label{tab:result_model_size}
\end{table}

\subsection{Excursion: Comparison with Triplet Extraction Results}
\label{sec:triples}
To evaluate the finer-grained aspect-based sentiment analysis triplet extraction task, we additionally extend an assessment of the target-aspect-sentiment detection (TASD) task using Qwen-2.5 7B and GPT-4o, which aims to extract sentiment triplets (aspect, category, sentiment), with the results reported in Table \ref{tab:triples_result}. We observe a performance decline across all five prompting strategies after adding the complexity of category recognition (including 12 categories in \S\ref{sec:detailed_aspect_categories}). For example, compared to the binary (aspect, sentiment) extraction task, the results on TASD task experience a 3.77\% drop in F1 score under the zero-shot strategy. This can be interpreted as the performance of the model deteriorating more noticeably with increasing task complexity. Therefore, it suggests that fine-grained multilingual sentiment triplet extraction tasks require more in-depth exploration and optimization of prompt strategies. Details of the prompts can be found in \S\ref{sec:Prompts}.

\begin{table}[htbp]
\centering

\begin{tabular}{c|c|cccccc}
\toprule
Methods&Model & EN & ES & FR & NL & RU & Avg. \\
\midrule

\multirow{2}{*}{Zero-Shot}&Qwen-2.5 7B&44.83& 41.88 &41.66 &  33.77 & 33.78 & 39.18 \\
&GPT-4o  &48.57& 43.28 &44.92  & 39.24 & 38.75 & 42.95 \\
\midrule
\multirow{2}{*}{CoT}&Qwen-2.5 7B&42.73& 38.28 &39.12 &  31.63 & 31.38 & 36.63 \\
&GPT-4o  &51.34& 48.89 &46.76  & 42.14 & 37.65 & 45.36 \\
\midrule
\multirow{2}{*}{Self-Improve.}&Qwen-2.5 7B&41.22& 36.57 &37.49 &  30.07 & 29.99 & 35.07 \\
&GPT-4o  &46.59& 42.37 &42.88  & 37.42 & 36.84 & 41.22 \\
\midrule
\multirow{2}{*}{Self-Debate}&Qwen-2.5 7B&37.23& 34.57 &34.96 &  28.17 & 26.16 & 32.22 \\
&GPT-4o  &44.23& 40.17 &41.22  & 35.84 & 34.98 & 39.29 \\
\midrule
\multirow{2}{*}{Self-Consist.}&Qwen-2.5 7B&43.53& 40.28 &40.31 &  32.55 & 32.47 & 37.83 \\
&GPT-4o  &49.33& 42.44 &43.79  & 38.14 & 36.95 & 42.13 \\
\bottomrule
\end{tabular}

\caption{The triplet extraction results with Qwen-2.5 7B and GPT-4o.}
\label{tab:triples_result}
\end{table}

\subsection{The Effect of Multi-Turn in Performance}
\label{sec:multi-diog}

\begin{figure}[htbp]
\centering
  \includegraphics[width=0.6\columnwidth]{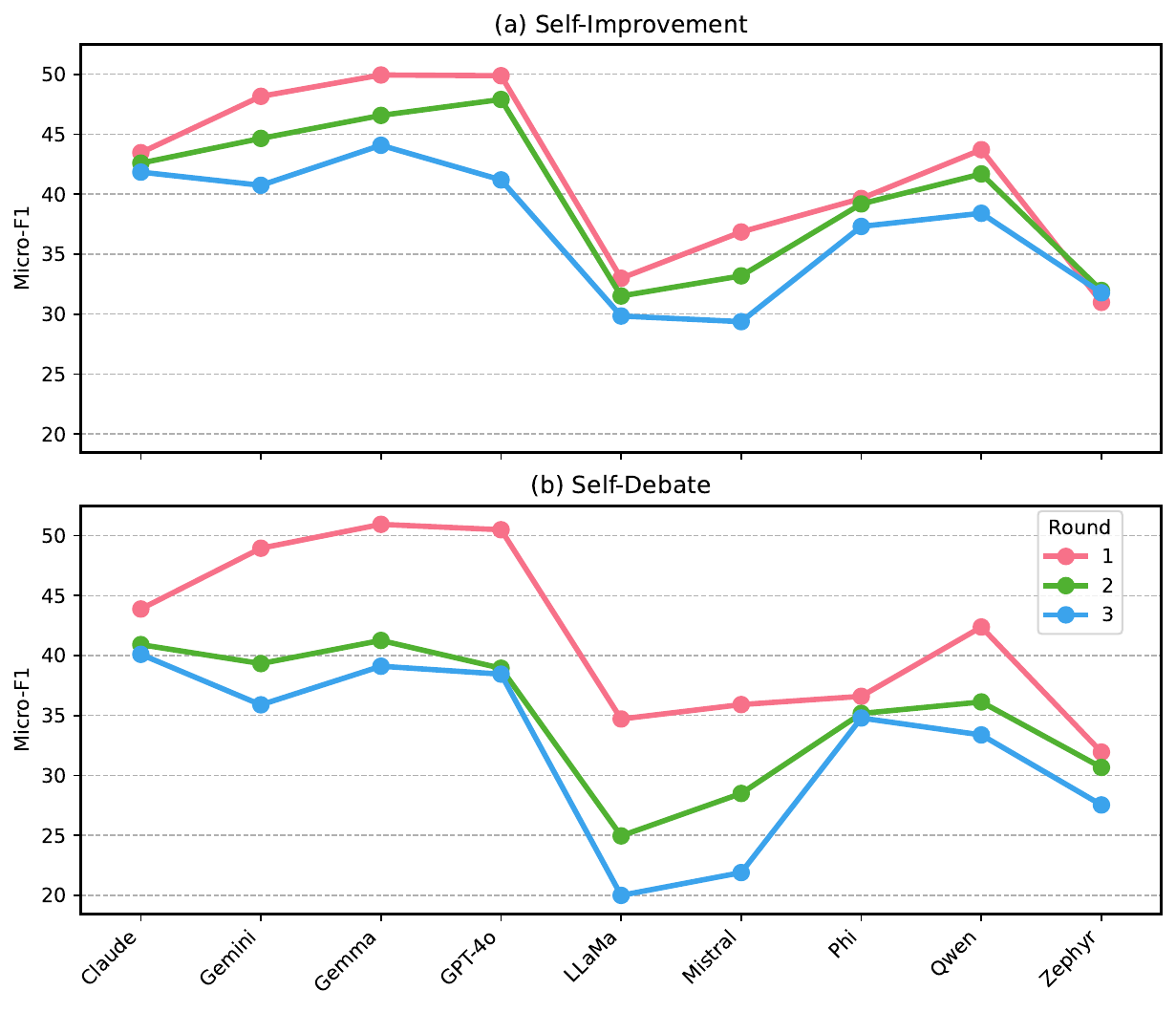}
  \caption{Multi-turn performance across different rounds in different models. In each model, The result is averaged across all the five languages.}
  \label{fig:improvement_debate}
\end{figure}

As discussed in \S\ref{sec:overall}, we observe that the performance of both multi-turn approaches, i.e. self-improvement and self-debate is worse than the zero-shot strategies. To better understand this outcome, we analyze the performance across different rounds. Figure \ref{fig:improvement_debate} presents the average performance differences across all languages in the three rounds for both multi-turn setups. Overall, the self-improvement setting yields higher average performance compared to the self-debate method, with an increase of 4.82\% in the three-round average F1 score. Furthermore, in each individual round, the self-improvement method achieves better performance compared to the self-debate method. Specifically, compared to the first round, the average F1 scores of the self-improvement and self-debate methods decreased by 1.84\% and 2.75\%, respectively, in the second round. Compared to the second round, the average F1 scores further decreased by 6.67\% and 2.74\%, respectively, in the third round. Notably, the performance degradation in the self-debate method is more pronounced across multiple dialogue rounds. Detailed performance for each language is provided in Appendix \ref{sec:result_details}.

Interestingly, for all LLMs, we find that performance consistently declines with each additional round. Notably, in the self-debate setup for GPT-4o, the performance between the first and second rounds drops by approximately 10\%. One possible explanation for this decline is that during the multi-turn process, when LLMs are prompted to self-correct, they tend to assume that errors exist in the prior responses, even when instructed to only revise incorrect predictions \citep{bai2024mt}. This assumption may lead to unnecessary changes and performance degradation.

\subsection{Zero-Shot Cross-Lingual Transfer with LLMs}
In addition, we evaluate the performance of fine-tuning the LLMs on the source language (English) and then validate on target languages, to assess the zero-shot cross-lingual transfer ability of LLMs via fine-tuning. We employ the LoRA approach \cite{hu2022lora}, and experiment on the open-source LLMs employed in the main experiments. For fine-tuning, we apply the base model of the LLMs. Table \ref{tab:lora} shows the detailed results of each language. The results of different LLMs vary, with Qwen2.5-7B achieving the highest average performance. In general, as expected, they perform better than the zero-shot prompting approaches. 

\begin{table}[ht]
\centering
\begin{tabular}{cccccc}
\toprule
& FR & ES & NL & RU & Avg \\
\midrule

Llama-3.1-8B + LoRA& 52.65 & 55.37 & 50.37 & 48.12 & 51.63 \\
Gemma-2-9B + LoRA & 48.17 & 57.46 & 51.97 & 47.65 & 51.31 \\
Mistral-7B-v0.3 + LoRA & 59.46 & 60.79 & 56.89 & 50.52 & 56.92 \\
Qwen2.5-7B + LoRA & 63.01 & 68.95 & 60.84 & 53.50 & 61.58 \\
Phi-3.5-mini + LoRA & 52.04 & 58.44 & 52.77 & 47.98 & 52.81 \\
Zephyr-7B + LoRA & 55.40 & 64.80 & 55.16 & 49.46 & 56.21 \\
\bottomrule
\end{tabular}
\caption{Performance of different LLMs in LoRA fine-tuning.}
\label{tab:lora}
\end{table}

Looking into the single languages, the performance across languages is inconsistent, with a noticeable drop in complex morphological languages such as Russian (53.50 for Qwen2.5-7B), in comparison with Spanish (68.95 for Qwen2.5-7B). This finding is consistent with the findings in the main experiments (\S\ref{sec:overall}) and in the supervised fine-tuning baseline results (\S\ref{sec:baselines}).

\subsection{Few-Shot Ablation}
In addition to the main prompting strategies described in \S\ref{sec:prompting}, we conduct few-shot experiments using the same setup as the zero-shot experiments. This allows us to examine how including examples in the prompt affects final performance.

In the few-shot setup, the model is provided with $n$ examples from the training data as contextual demonstrations. A KNN-based retrieval strategy is employed, selecting the top-$n$ nearest examples based on the similarity to the input sentence, as computed using Sentence Transformers \cite{reimers-gurevych-2019-sentence}. The top $n$ most similar sentences are included as examples, with experiments conducted using the following values for $n$: $n \in [1, 2, 4, 8, 16]$.

\begin{figure}[htbp]
\centering
  \includegraphics[width=0.6\columnwidth]{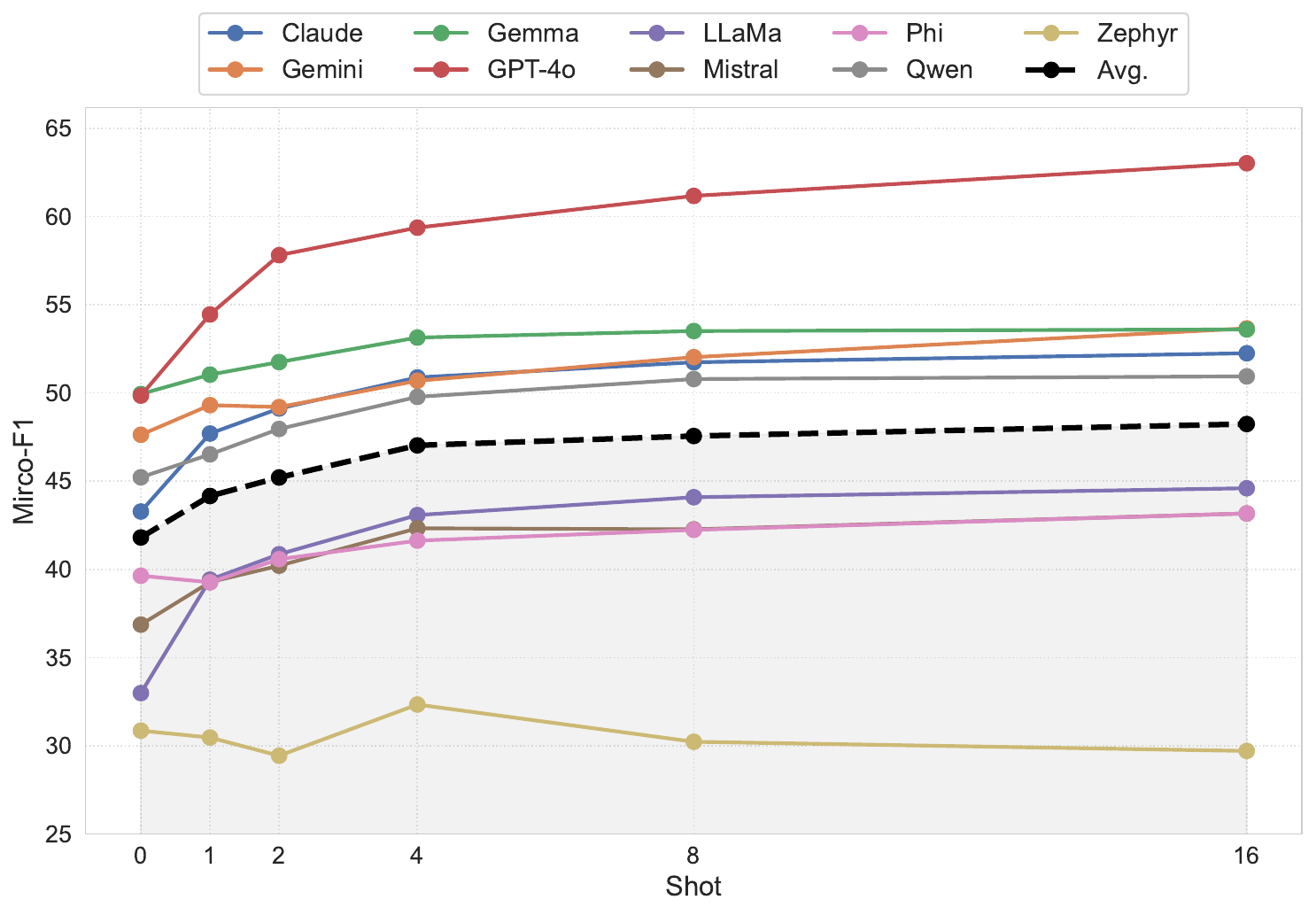}
  \caption{Few-shot results across different shots. In each shot, the result is averaged across all five languages.}
  \label{fig:fewshot}
\end{figure}

Figure \ref{fig:fewshot} shows the results of the few-shot experiments across different shot sizes for all LLMs. We observe that in most cases, as the number of examples increases, model performance improves accordingly. This supports the assumption that including semantically related examples in the prompt enhances the LLMs' inference capabilities in ABSA task. The exception is Zephyr, whose performance remains unstable and consistently lower than other models, as also noted in the main experiment results in all different setups (see Table \ref{tab:performance}). Detailed results for the few-shot experiments are presented in Table \ref{tab:fewshot} in the Appendix.

\subsection{Error Cases} In this section, we analyze the types of errors observed during the experiments. To classify the errors, we follow the categorization proposed by \cite{xie-etal-2023-empirical}, which also addresses token-level tasks like NER. We categorize the prediction errors in sentiment analysis into two main types: aspect term boundaries and sentiment polarity. The error types related to aspect term boundaries include Contain gold, Contained by gold, Overlap with gold, Completely-O, and Omitted mentions. The error types related to sentiment polarity include Out-of-Domain (OOD) types and Wrong types, as shown in the figure below:

\begin{tcolorbox}
    [colback=gray!15, colframe=gray!100, sharp corners, leftrule={2pt}, rightrule={0pt}, toprule={0pt}, bottomrule={0pt}, left={2pt}, right={2pt}, top={3pt}, bottom={3pt}]
\begin{itemize}[leftmargin=*,nolistsep] 
\scriptsize
\item \textbf{Out-of-Domain (OOD) types}: Predicted sentiment types that are not in the predefined label set. 
\item \textbf{Wrong types}: Predicted sentiment types that are incorrect but still belong to the label set. 
\item \textbf{Contain gold}: Predicted aspects that include the correct (gold) aspects. 
\item \textbf{Contained by gold}: Predicted aspects that are fully contained within the gold aspects. 
\item \textbf{Overlap with gold}: Predicted aspects that overlap with gold aspects but do not fall into the two previous categories. 
\item \textbf{Completely-O}: Predicted aspects that are entirely outside the gold aspects, yet overlap with other tokens. 
\item \textbf{OOD mentions}: Predicted aspects that do not appear in input text at all. 
\item \textbf{Omitted mentions}: Aspects that should have been predicted but were omitted by the model.
\end{itemize}
\label{tab:errortype}
\end{tcolorbox}

\begin{figure*}
    \centering
    \includegraphics[width=1\linewidth]{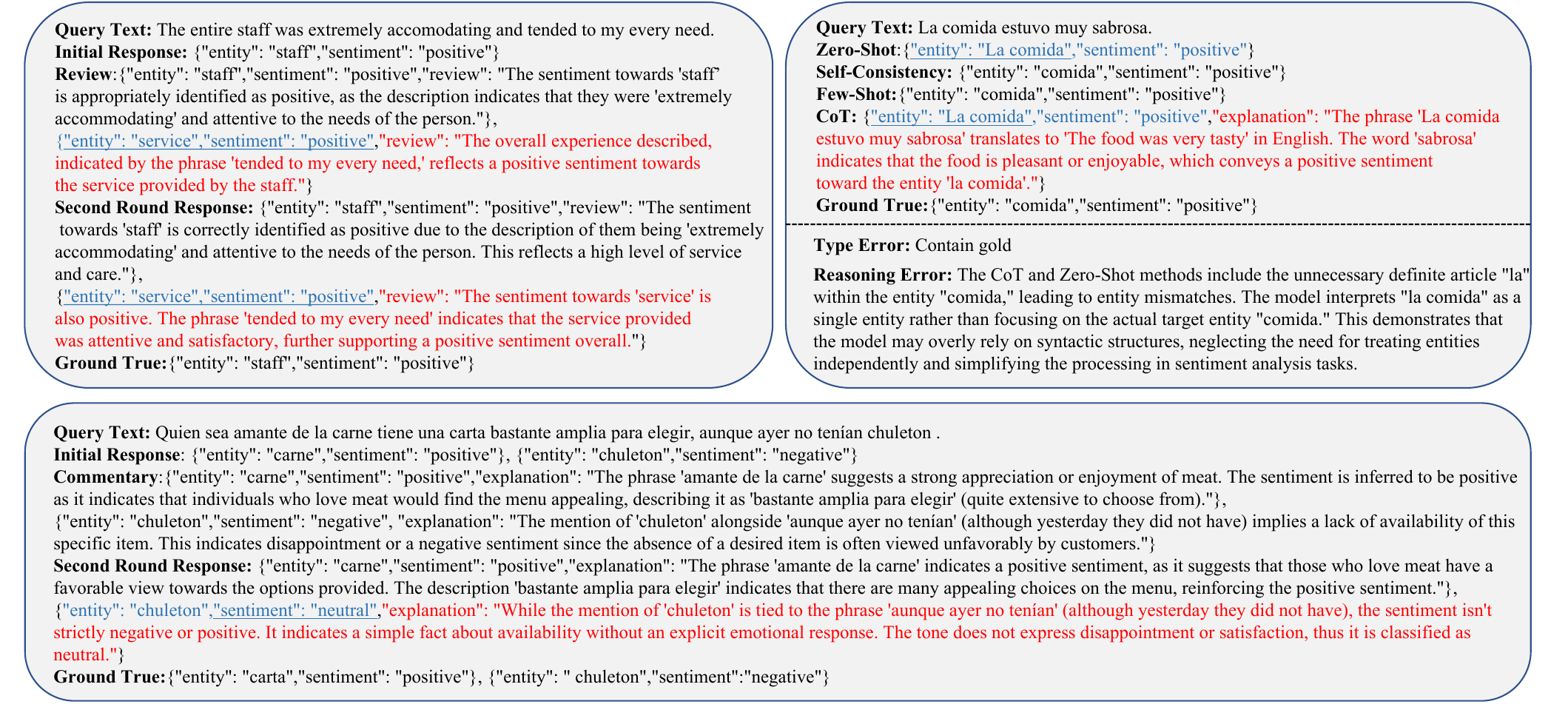}
    \caption{Case study of error increase with the proposed methods. Texts in blue are involved entities in the error cases. The items highlighted are false predictions.}
    \label{fig:error_case3}
\end{figure*}

\begin{figure}[htbp]
\centering
  \includegraphics[width=0.6\columnwidth]{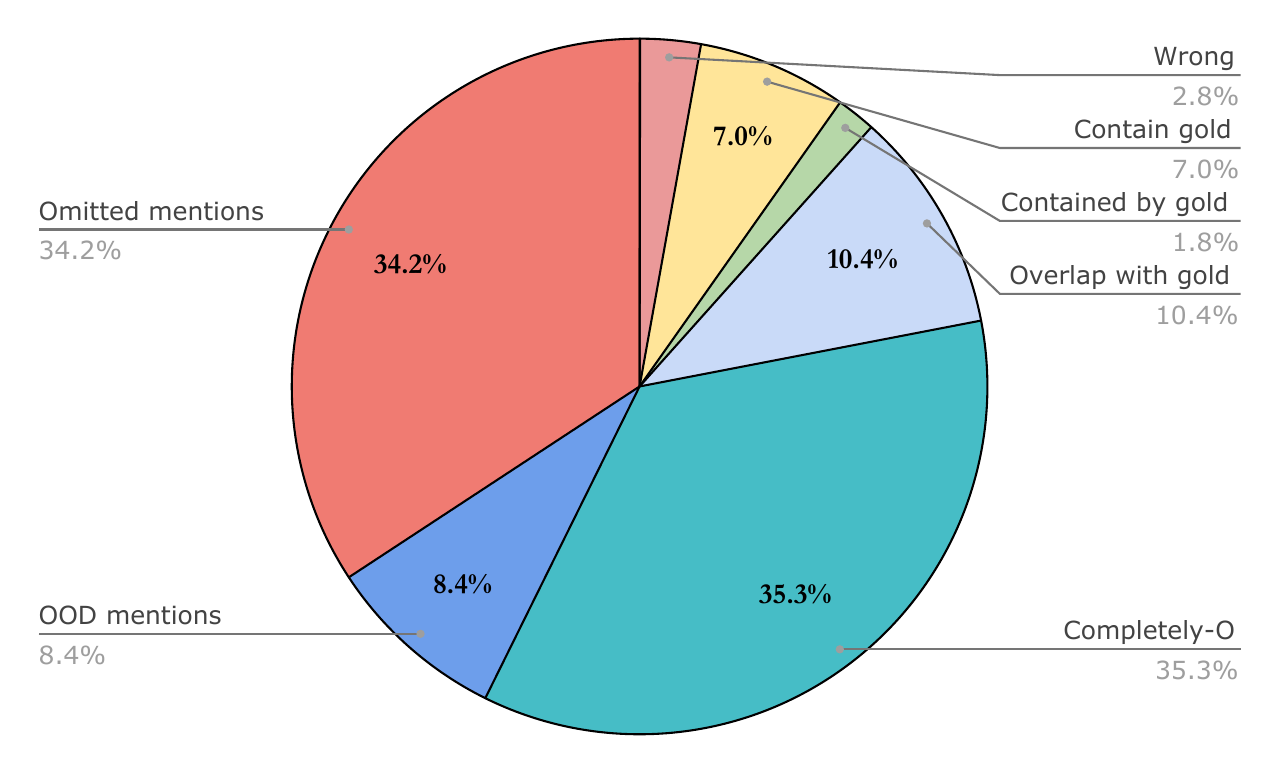}
  \caption{Percentage of different error types with GPT-4o under the zero-shot prompting strategy. The percentages are averaged across the five languages.}
  \label{fig:error}
\end{figure}

We use the zero-shot performance of the GPT-4o model as a representative example. Figure \ref{fig:error} illustrates the distribution of error types for GPT-4o under the zero-shot setting. The most frequent error types are “Completely-O” and “Omitted mentions”, together accounting for over 70\% of the total errors. This observation is partially consistent with the findings of \cite{xie-etal-2023-empirical} on NER tasks, where “Completely-O” and “Wrong types" are the most common error types. The “Completely-O" errors may result from incomplete annotations or the model's tendency to guess aspect terms based on prior knowledge. The “Omitted mentions" errors likely stem from inadequate contextual understanding.

As shown in Figure \ref{fig:error_case3}, we select three typical cases to illustrate reasoning errors, which include OOD mentions, contain gold, and wrong types. In Case 1, the aspect “service" is over-predicted during the first review, which subsequently leads to an incorrect final prediction. Moreover, despite providing a specific task description before each round of debate, the model focuses on the sentiment polarity of the entire sentence rather than the specific aspect terms. Consequently, the entire sentence is generalized as “service is highly satisfactory", resulting in an OOD prediction. We hypothesize that the model suffers from catastrophic forgetting during complex contextual reasoning. In Case 2, the CoT method first translates the original sentence back into English before making predictions, focusing heavily on analyzing the sentence's structural features. This leads to an incorrect identification of “La comida" as an aspect term. We can explain this as the CoT method places too much emphasis on syntactic structure analysis rather than accurately identifying aspect boundaries. In Case 3, after two rounds of self-improvement corrections, the model incorrectly revises a previously correct prediction, changing the sentiment polarity of “chuleton" to neutral. We suspect that the model encounters ambiguity in the implicit sentiment reasoning process for the aspect “chuleton" and lacks confidence during self-correction, ultimately leading to an incorrect prediction type.

Overall, the errors in the selected cases mentioned can be attributed to the following reasons: neglecting the reasoning of entity boundaries during the inference process, error accumulation from multiple rounds of dialogue, and ambiguity in implicit sentiment reasoning.

\section{Conclusion}
In this study, we conduct an empirical evaluation of zero-shot multilingual ABSA using different LLMs, exploring their performance across five different prompting methods. Inspired by LLMs' strong reasoning capabilities in logical and arithmetic tasks, we apply these strategies and assess their effectiveness across five languages: English, French, Spanish, Dutch, and Russian, as well as domain-specific datasets. Our findings indicate that LLMs do not surpass fully-supervised multilingual models in the ABSA task. Performance varies significantly across languages, with higher-resource languages like English yielding better results. We also find that closed-source models generally outperform open-source ones, and simpler zero-shot prompts prove more effective than more complex prompting strategies. These insights suggest that while LLMs show potential for multilingual tasks, their effectiveness remains highly dependent on the task, language, and model configuration, highlighting the need for further exploration to optimize LLMs for cross-lingual tasks like ABSA.

\bmhead{Acknowledgements}

The authors would like to respect and thank all reviewers for their constructive and helpful review.

\section*{Declarations}

\begin{itemize}
\item \textbf{Funding.} This work was supported in part by the Guangdong Basic and Applied Basic Research Foundation under Grant 2023A1515011370, the National Natural Science Foundation of China (32371114), the Characteristic Innovation Projects of Guangdong Colleges and Universities (No. 2018KTSCX049), the Guangdong Provincial Key Laboratory [No.2023B1212060076].
\item \textbf{Conflict of interest.} The authors declare no competing interests.
\item \textbf{Ethics approval.} This atricle does not contain any studies with human participants or animals performed by any of the authors.

\item \textbf{Data availability.} Data sharing is not applicable to this article as no dataset was generated or analyzed during the current study. The dataset used in this paper is public and available.

\end{itemize}

\begin{appendices}

\section{Details of Prompts and Experiments}
\label{sec:appendix}

\subsection{Prompt}
\label{sec:Prompts}
We show all of our prompts for six methods. The prompts are in Figure \ref{fig:single-turn-prompt}-\ref{fig:self-debate-prompt-triplet}.

\subsection{Detailed Results}
\label{sec:result_details}
The details of the overall performance under the self-improvement, self-debate, and few-shot settings are shown in Table \ref{tab:improve}, \ref{tab:debate} and \ref{tab:fewshot}.

\subsection{Fine-Tuning Parameter Setting}
\label{sec:ft_parameter}
The details of the parameter setting under supervised methods are shown in Table \ref{tab:hyperparameters}.

\subsection{Dataset Details}
\label{sec:dataset}
The details of the dataset statistics in each language are shown in Table \ref{tab:dataset-semeval}. 
\# S and \# A denote the number of sentences and aspect terms in different sets, respectively.

\begin{table}[htbp]
\centering
\small
\renewcommand\arraystretch{1}
\begin{tabular}{lc}
\toprule
\textbf{Parameter}          & \textbf{Value}     \\ \midrule
Epoch                       & 20           \\ 
Batch size                  & [8, 16]            \\ 
Learning rate (BERT)        & 5e-5      \\ 
Learning rate (XLM-R)       & 2e-5              \\ 
Hidden size         & 768                \\ 
Dropout rate               & 0.1                \\ 
Max steps & [2000,2500]                 \\ 
Adam epsilon                & 1e-8               \\ 
Warm factor                 & 0.1                \\ \bottomrule
\end{tabular}
\caption{Fine-tuning hyper-parameter settings.}
\label{tab:hyperparameters}
\end{table}

\begin{table}[htbp]
\centering
\small
\renewcommand\arraystretch{1}
\setlength\tabcolsep{4pt}
\begin{tabular}{lcccccc}
\toprule
          &  &EN&FR&ES&NL&RU    \\ \midrule
\multirow{2}{*}{\textbf{Train} }           & \# S          & 2000 &1664 &2070 &1722 &3655           \\ 
         & \# A     &1743 &1641 &1856 &1231 &3077       \\ \midrule
\multirow{2}{*}{\textbf{Test}}            &\# S     &676 &668 &881 &575& 1209       \\ 
          &\# A      &612 &650 &713 &373 &949       \\ 
\bottomrule
\end{tabular}
\caption{Statistic of the original dataset.}
\label{tab:dataset-semeval}
\end{table}

\subsection{Detailed Aspect Categories}
\label{sec:detailed_aspect_categories}
We present the detailed aspect categories for the dataset in the previous section (\S\ref{sec:triples}).
\begin{tcolorbox}
    [colframe=gray!100, sharp corners, leftrule={3pt}, rightrule={0pt}, toprule={0pt}, bottomrule={0pt}, left={2pt}, right={2pt}, top={3pt}, bottom={3pt}]
    \small
\textbf{Category} =  ['FOOD\#QUALITY',
'AMBIENCE\#GENERAL',
'LOCATION\#GENERAL',
'SERVICE\#GENERAL',
'RESTAURANT\#GENERAL',
'DRINKS\#QUALITY',
'RESTAURANT\#MISCELLANEOUS',
'DRINKS\#PRICES',
'DRINKS\#STYLE\_OPTIONS',
'FOOD\#PRICES',
'FOOD\#STYLE\_OPTIONS',
'RESTAURANT\#PRICES']
\end{tcolorbox}

\begin{figure*}
\centering
\begin{tcolorbox}[colframe=gray!50!black, colback=gray!5!white, title=Zero-Shot \& Self-Consistency ]
\scriptsize

Aspect-Based Sentiment Analysis (ABSA) involves identifying specific entity (such as a person, product, service, or experience) mentioned in a text and determining the sentiment expressed toward each entity. Each entity is associated with a sentiment that can be [positive, negative, or neutral].

\vspace{3pt}
Your task is to:

\vspace{3pt}
1. Identify the entity with a sentiment mentioned in the given text.\\
2. For each identified entity, determine the sentiment in the label set (positive, negative, or neutral).\\
3. The output should be a list of dictionaries, where each dictionary contains the entity with a sentiment and its corresponding sentiment. If there are no sentiment-bearing entities in the text, the output should be an empty list.

\vspace{3pt}
Example Output format:

\vspace{3pt}
[{“entity": “\textless entity\textgreater", “sentiment": “\textless label\textgreater"}]

\vspace{3pt}
Please return the final output based on the following text in json format.

\end{tcolorbox}

\begin{tcolorbox}[colframe=gray!50!black, colback=gray!5!white, title=CoT ]
\scriptsize

Aspect-Based Sentiment Analysis (ABSA) involves identifying specific entity (such as a person, product, service, or experience) mentioned in a text and determining the sentiment expressed toward each entity. Each entity is associated with a sentiment that can be [positive, negative, or neutral]. 

\vspace{3pt}
Your task is to:

\vspace{3pt}
1. Identify the entity with a sentiment mentioned in the given text. 

2. For each identified entity, determine the sentiment in the label set (positive, negative, or neutral).

3. Provide a reasoning process for how you identified the entities and assigned their sentiments.

4. The output should be a list of dictionaries, where each dictionary contains the entity with a sentiment and its reasoning process. If there are no sentiment-bearing entities in the text, the output should be an empty list.

\vspace{3pt}
Example Output format:

\vspace{3pt}
[{“entity": “\textless entity\textgreater", “sentiment": “\textless label\textgreater", “Explanation": “\textless reasoning process\textgreater"}]

\vspace{3pt}
Please return the final output base on the following text in json format.
\end{tcolorbox}

\begin{tcolorbox}[colframe=gray!50!black, colback=gray!5!white, title=Few-Shot ]
\scriptsize

Aspect-Based Sentiment Analysis (ABSA) involves identifying specific entity (such as a person, product, service, or experience) mentioned in a text and determining the sentiment expressed toward each entity.

\vspace{3pt}
Each entity is associated with a sentiment that can be [positive, negative, or neutral].

\vspace{3pt}
Your task is to:

\vspace{3pt}
1. Identify the entity with a sentiment mentioned in the given text.

2. For each identified entity, determine the sentiment in the label set (positive, negative, or neutral).

3. The output should be a list of dictionaries, where each dictionary contains the entity with a sentiment and its corresponding sentiment. If there are no sentiment-bearing entities in the text, the output should be an empty list.

\vspace{3pt}
Example Output format:

\vspace{3pt}
[{“entity": “\textless entity\textgreater", “sentiment": “\textless label\textgreater"}]

\vspace{3pt}
Here are \{top\_k\} similar sentences from the training set:

\vspace{3pt}
\{similar\_sentences and corresponding entity with its sentiment\}

\vspace{3pt}
Please return the final output based on the following text in json format.
\end{tcolorbox}

\caption{Single-turn prompts used in this paper for tuple extraction.}
\label{fig:single-turn-prompt}
\end{figure*}

\begin{figure*}
\centering
\begin{tcolorbox}[colframe=gray!50!black, colback=gray!5!white, title=Zero-Shot \& Self-Consistency ]
\scriptsize

Aspect-Based Sentiment Analysis (ABSA) requires identifying specific entities mentioned in a text and determining the sentiment expressed toward each entity.\\ Each entity is associated with:
\\ A category from the list: [FOOD\#QUALITY,
AMBIENCE\#GENERAL,
LOCATION\#GENERAL,
SERVICE\#GENERAL,
RESTAURANT\#GENERAL,
DRINKS\#QUALITY,
RESTAURANT\#MISCELLANEOUS,
DRINKS\#PRICES,
DRINKS\#STYLE\_OPTIONS,
FOOD\#PRICES,
FOOD\#STYLE\_OPTIONS,
RESTAURANT\#PRICES]

A sentiment: [positive, negative, neutral].

\vspace{3pt}
Your task is to:

\vspace{3pt}
1. Identify entities in the text, along with their categories and sentiments.\\
2. For each identified entity, assign a category from the provided category list.\\
3. Determine the sentiment for each entity as one of [positive, negative, neutral].\\
4. Return the results as a list of dictionaries, each containing the entity, category, and sentiment. If no entities are found, return an empty list.\\

\vspace{3pt}
Example Output format:

\vspace{3pt}
[{“entity": “\textless entity\textgreater", “category": “\textless category\textgreater", “sentiment": “\textless label\textgreater"}]

\vspace{3pt}
Please return the final output based on the following text in json format.

\end{tcolorbox}

\begin{tcolorbox}[colframe=gray!50!black, colback=gray!5!white, title=CoT ]
\scriptsize

Aspect-Based Sentiment Analysis (ABSA) requires identifying specific entities mentioned in a text and determining the sentiment expressed toward each entity.\\ Each entity is associated with:
\\ A category from the list: [FOOD\#QUALITY,
AMBIENCE\#GENERAL,
LOCATION\#GENERAL,
SERVICE\#GENERAL,
RESTAURANT\#GENERAL,
DRINKS\#QUALITY,
RESTAURANT\#MISCELLANEOUS,
DRINKS\#PRICES,
DRINKS\#STYLE\_OPTIONS,
FOOD\#PRICES,
FOOD\#STYLE\_OPTIONS,
RESTAURANT\#PRICES]

A sentiment: [positive, negative, neutral].

\vspace{3pt}
Your task is to:

\vspace{3pt}



1. Identify entities in the text, along with their categories and sentiments.\\
2. For each identified entity, assign a category from the provided category list.\\
3. Determine the sentiment for each entity as one of [positive, negative, neutral].\\
4. Provide a reasoning process for how you identified the entities, categories, and sentiments.\\
5. Return the results as a list of dictionaries, each containing the entity, category, and sentiment. If no entities are found, return an empty list.\\

\vspace{3pt}
Example Output format:

\vspace{3pt}
[{“entity": “\textless entity\textgreater", “category": “\textless category\textgreater", “sentiment": “\textless label\textgreater", “Explanation": “\textless reasoning process\textgreater"}]

\vspace{3pt}
Please return the final output based on the following text in json format.
\end{tcolorbox}

\caption{Single-turn prompts used in this paper for triplet extraction.}
\label{fig:single-turn-prompt-triple}
\end{figure*}

\newpage

\begin{figure*}
\centering
\begin{tcolorbox}[colframe=gray!50!black, colback=gray!5!white, title=Round 1]
\scriptsize
\vspace{5pt}

Aspect-Based Sentiment Analysis (ABSA) involves identifying specific entity (such as a person, product, service, or experience) mentioned in a text and determining the sentiment expressed toward each entity. Each entity is associated with a sentiment that can be [positive, negative, or neutral].

\vspace{3pt}
Your task is to:

\vspace{3pt}
1. Identify the entity with a sentiment mentioned in the given text.\\
2. For each identified entity, determine the sentiment in the label set (positive, negative, or neutral).\\
3. The output should be a list of dictionaries, where each dictionary contains the entity with a sentiment and its corresponding sentiment. If there are no sentiment-bearing entities in the text, the output should be an empty list.

\vspace{3pt}
Example Output format:

\vspace{3pt}
[{“entity": “\textless entity\textgreater", “sentiment": “\textless label\textgreater"}]

\vspace{3pt}
Please return the final output based on the following text in json format.

\end{tcolorbox}

\begin{tcolorbox}[colframe=gray!50!black, colback=gray!5!white, title=Round 2]
\scriptsize

\vspace{5pt}
Aspect-Based Sentiment Analysis (ABSA) involves identifying specific entity (such as a person, product, service, or experience) mentioned in a text and determining the sentiment expressed toward each entity. Each entity is associated with a sentiment that can be [positive, negative, or neutral].

\vspace{3pt}
Here is the sentence: \{sentence\}. You have classified the sentiment of the entities in this sentence.

\vspace{3pt}
Here is your initial result: \{initial\_result\}. 

\vspace{3pt}
Please explain why you classified them in this way.

\vspace{3pt}
Example Output format:

\vspace{3pt}
[{“entity": “\textless entity\textgreater", “sentiment": “\textless label\textgreater", “Explanation": “\textless reasoning process\textgreater"}]

\vspace{3pt}
Please return the final output based on the above sentence in json format.

\end{tcolorbox}

\begin{tcolorbox}[colframe=gray!50!black, colback=gray!5!white, title=Round 3]
\scriptsize

\vspace{5pt}
Aspect-Based Sentiment Analysis (ABSA) involves identifying specific entity (such as a person, product, service, or experience) mentioned in a text and determining the sentiment expressed toward each entity.

\vspace{3pt}
Each entity is associated with a sentiment that can be [positive, negative, or neutral].

\vspace{3pt}
Here is the sentence: \{sentence\}. You have given a sentiment classification for the entities in this sentence.

\vspace{3pt}
Here is your initial result: \{initial\_result\}. 

\vspace{3pt}
Here is your explanation: \{explanation\_result\}.

\vspace{3pt}
Please recheck your classification and explanation. If you find any errors or better classifications, please update your response accordingly. 

\vspace{3pt}
Example Output format:

\vspace{3pt}
[{“entity": “\textless entity\textgreater", “sentiment": “\textless label\textgreater", “Explanation": “\textless reasoning process\textgreater"}]

\vspace{3pt}
Please return the final output based on the above sentence in json format.

\end{tcolorbox}
\caption{Self-improvement prompts used in this paper for tuple extraction.}
\label{fig:self-improvement-prompt}
\end{figure*}

\begin{figure*}
\centering
\begin{tcolorbox}[colframe=gray!50!black, colback=gray!5!white, title=Round 1]
\scriptsize
\vspace{5pt}

Aspect-Based Sentiment Analysis (ABSA) requires identifying specific entities mentioned in a text and determining the sentiment expressed toward each entity.\\ Each entity is associated with:
\\ A category from the list: [FOOD\#QUALITY,
AMBIENCE\#GENERAL,
LOCATION\#GENERAL,
SERVICE\#GENERAL,
RESTAURANT\#GENERAL,
DRINKS\#QUALITY,
RESTAURANT\#MISCELLANEOUS,
DRINKS\#PRICES,
DRINKS\#STYLE\_OPTIONS,
FOOD\#PRICES,
FOOD\#STYLE\_OPTIONS,
RESTAURANT\#PRICES]

A sentiment: [positive, negative, neutral].

\vspace{3pt}
Your task is to:

\vspace{3pt}
1. Identify entities in the text, along with their categories and sentiments.\\
2. For each identified entity, assign a category from the provided category list.\\
3. Determine the sentiment for each entity as one of [positive, negative, neutral].\\
4. Return the results as a list of dictionaries, each containing the entity, category, and sentiment. If no entities are found, return an empty list.\\

\vspace{3pt}
Example Output format:

\vspace{3pt}
[{“entity": “\textless entity\textgreater", “category": “\textless category\textgreater", “sentiment": “\textless label\textgreater"}]

\vspace{3pt}
Please return the final output based on the following text in json format.

\end{tcolorbox}

\begin{tcolorbox}[colframe=gray!50!black, colback=gray!5!white, title=Round 2]
\scriptsize

\vspace{5pt}
Aspect-Based Sentiment Analysis (ABSA) requires identifying specific entities mentioned in a text and determining the sentiment expressed toward each entity.\\ Each entity is associated with:
\\ A category from the list: [FOOD\#QUALITY,
AMBIENCE\#GENERAL,
LOCATION\#GENERAL,
SERVICE\#GENERAL,
RESTAURANT\#GENERAL,
DRINKS\#QUALITY,
RESTAURANT\#MISCELLANEOUS,
DRINKS\#PRICES,
DRINKS\#STYLE\_OPTIONS,
FOOD\#PRICES,
FOOD\#STYLE\_OPTIONS,
RESTAURANT\#PRICES]

A sentiment: [positive, negative, neutral].

\vspace{3pt}
Here is the sentence: \{sentence\}. You have classified the sentiment and categories of the entities in this sentence.

\vspace{3pt}
Here is your initial result: \{initial\_result\}. 

\vspace{3pt}
Please explain why you classified them in this way.

\vspace{3pt}
Example Output format:

\vspace{3pt}
[{“entity": “\textless entity\textgreater", “category": “\textless category\textgreater", “sentiment": “\textless label\textgreater", “Explanation": “\textless reasoning process\textgreater"}]

\vspace{3pt}
Please return the final output based on the above sentence in json format.

\end{tcolorbox}

\begin{tcolorbox}[colframe=gray!50!black, colback=gray!5!white, title=Round 3]
\scriptsize

\vspace{5pt}
Aspect-Based Sentiment Analysis (ABSA) requires identifying specific entities mentioned in a text and determining the sentiment expressed toward each entity.\\ Each entity is associated with:
\\ A category from the list: [FOOD\#QUALITY,
AMBIENCE\#GENERAL,
LOCATION\#GENERAL,
SERVICE\#GENERAL,
RESTAURANT\#GENERAL,
DRINKS\#QUALITY,
RESTAURANT\#MISCELLANEOUS,
DRINKS\#PRICES,
DRINKS\#STYLE\_OPTIONS,
FOOD\#PRICES,
FOOD\#STYLE\_OPTIONS,
RESTAURANT\#PRICES]

A sentiment: [positive, negative, neutral].


\vspace{3pt}
Here is the sentence: \{sentence\}. You have given a sentiment and category classification for the entities in this sentence.

\vspace{3pt}
Here is your initial result: \{initial\_result\}. 

\vspace{3pt}
Here is your explanation: \{explanation\_result\}.

\vspace{3pt}
Please recheck your classification and explanation. If you find any errors or better classifications, please update your response accordingly. 

\vspace{3pt}
Example Output format:

\vspace{3pt}
[{“entity": “\textless entity\textgreater", “category": “\textless category\textgreater", “sentiment": “\textless label\textgreater", “Explanation": “\textless reasoning process\textgreater"}]

\vspace{3pt}
Please return the final output based on the above sentence in json format.

\end{tcolorbox}
\caption{Self-improvement prompts used in this paper for triplet extraction.}
\label{fig:self-improvement-prompt-triplet}
\end{figure*}

\newpage

\begin{figure*}
\centering
\begin{tcolorbox}[colframe=gray!50!black, colback=gray!5!white, title=Round 1]
\scriptsize
\vspace{5pt}

Aspect-Based Sentiment Analysis (ABSA) involves identifying specific entity (such as a person, product, service, or experience) mentioned in a text and determining the sentiment expressed toward each entity. Each entity is associated with a sentiment that can be [positive, negative, or neutral].

\vspace{3pt}
Your task is to:

\vspace{3pt}
1. Identify the entity with a sentiment mentioned in the given text. 

2. For each identified entity, determine the sentiment in the label set (positive, negative, or neutral).

3. The output should be a list of dictionaries, where each dictionary contains the entity with a sentiment and its corresponding sentiment. If there are no sentiment-bearing entities in the text, the output should be an empty list.

\vspace{3pt}
Example Output format:

\vspace{3pt}
Round 1: [{“entity": “\textless entity\textgreater", “sentiment": “\textless label\textgreater"}]

\vspace{3pt}
Please return the final output based on the following text in json format.

\end{tcolorbox}

\begin{tcolorbox}[colframe=gray!50!black, colback=gray!5!white, title=Round 2]
\scriptsize

\vspace{5pt}
Aspect-Based Sentiment Analysis (ABSA) involves identifying specific entity (such as a person, product, service, or experience) mentioned in a text and determining the sentiment expressed toward each entity. Each entity is associated with a sentiment that can be [positive, negative, or neutral].

\vspace{3pt}
The source sentence is: \{sentence\}. 

\vspace{3pt}
The first response result: \{previous\_response\}

\vspace{3pt}
Please review and comment on the following response. Provide corrections if necessary or add more details to improve the result.

\vspace{3pt}
Example Output format:

\vspace{3pt}
Round 2: [{“entity": “\textless entity\textgreater", “sentiment": “\textless label\textgreater",“Review":“\textless review\textgreater"}]

\vspace{3pt}
Please return the final output based on the above sentence in json format.

\end{tcolorbox}

\begin{tcolorbox}[colframe=gray!50!black, colback=gray!5!white, title=Round 3]
\scriptsize

\vspace{5pt}
Aspect-Based Sentiment Analysis (ABSA) involves identifying specific entity (such as a person, product, service, or experience) mentioned in a text and determining the sentiment expressed toward each entity. Each entity is associated with a sentiment that can be [positive, negative, or neutral].

\vspace{3pt}
The source sentence is: \{sentence\}. 

\vspace{3pt}
The first response result: \{first\_response\}.

\vspace{3pt}
The first commentary result: \{first\_commentary\}.

\vspace{3pt}
Based on the initial response and commentary, please further debate and refine the analysis. If there are any conflicting opinions or uncertainties, resolve them (both entity and review) and provide a more detailed and accurate response.

\vspace{3pt}
Example Output format:

\vspace{3pt}
Round 3: [{“entity": “\textless entity\textgreater", “sentiment": “\textless label\textgreater",“Review":“\textless review\textgreater"}]

\vspace{3pt}
Please return the final output based on the above sentence in json format.

\vspace{5pt}
\end{tcolorbox}
\caption{Self-debate prompts used in this paper for tuple extraction.}
\label{fig:self-debate-prompt}
\end{figure*}

\begin{figure*}
\centering
\begin{tcolorbox}[colframe=gray!50!black, colback=gray!5!white, title=Round 1]
\scriptsize
\vspace{5pt}

Aspect-Based Sentiment Analysis (ABSA) requires identifying specific entities mentioned in a text and determining the sentiment expressed toward each entity.\\ Each entity is associated with:
\\ A category from the list: [FOOD\#QUALITY,
AMBIENCE\#GENERAL,
LOCATION\#GENERAL,
SERVICE\#GENERAL,
RESTAURANT\#GENERAL,
DRINKS\#QUALITY,
RESTAURANT\#MISCELLANEOUS,
DRINKS\#PRICES,
DRINKS\#STYLE\_OPTIONS,
FOOD\#PRICES,
FOOD\#STYLE\_OPTIONS,
RESTAURANT\#PRICES]

A sentiment: [positive, negative, neutral].

\vspace{3pt}
Your task is to:

\vspace{3pt}
1. Identify the entity with a sentiment mentioned in the given text. 

2. For each identified entity, determine the sentiment in the label set (positive, negative, or neutral).

3. The output should be a list of dictionaries, where each dictionary contains the entity with a sentiment and its corresponding sentiment. If there are no sentiment-bearing entities in the text, the output should be an empty list.

\vspace{3pt}
Example Output format:

\vspace{3pt}
Round 1: [{“entity": “\textless entity\textgreater", “category": “\textless category\textgreater", “sentiment": “\textless label\textgreater"}]

\vspace{3pt}
Please return the final output based on the following text in json format.

\end{tcolorbox}

\begin{tcolorbox}[colframe=gray!50!black, colback=gray!5!white, title=Round 2]
\scriptsize

\vspace{5pt}
Aspect-Based Sentiment Analysis (ABSA) requires identifying specific entities mentioned in a text and determining the sentiment expressed toward each entity.\\ Each entity is associated with:
\\ A category from the list: [FOOD\#QUALITY,
AMBIENCE\#GENERAL,
LOCATION\#GENERAL,
SERVICE\#GENERAL,
RESTAURANT\#GENERAL,
DRINKS\#QUALITY,
RESTAURANT\#MISCELLANEOUS,
DRINKS\#PRICES,
DRINKS\#STYLE\_OPTIONS,
FOOD\#PRICES,
FOOD\#STYLE\_OPTIONS,
RESTAURANT\#PRICES]

A sentiment: [positive, negative, neutral].

\vspace{3pt}
The source sentence is: \{sentence\}. 

\vspace{3pt}
The first response result: \{previous\_response\}

\vspace{3pt}
Please review and comment on the following response. Provide corrections if necessary or add more details to improve the result.

\vspace{3pt}
Example Output format:

\vspace{3pt}
Round 2: [{“entity": “\textless entity\textgreater", “category": “\textless category\textgreater", “sentiment": “\textless label\textgreater",“Review":“\textless review\textgreater"}]

\vspace{3pt}
Please return the final output based on the above sentence in json format.

\end{tcolorbox}

\begin{tcolorbox}[colframe=gray!50!black, colback=gray!5!white, title=Round 3]
\scriptsize

\vspace{5pt}
Aspect-Based Sentiment Analysis (ABSA) requires identifying specific entities mentioned in a text and determining the sentiment expressed toward each entity.\\ Each entity is associated with:
\\ A category from the list: [FOOD\#QUALITY,
AMBIENCE\#GENERAL,
LOCATION\#GENERAL,
SERVICE\#GENERAL,
RESTAURANT\#GENERAL,
DRINKS\#QUALITY,
RESTAURANT\#MISCELLANEOUS,
DRINKS\#PRICES,
DRINKS\#STYLE\_OPTIONS,
FOOD\#PRICES,
FOOD\#STYLE\_OPTIONS,
RESTAURANT\#PRICES]

A sentiment: [positive, negative, neutral].

\vspace{3pt}
The source sentence is: \{sentence\}. 

\vspace{3pt}
The first response result: \{first\_response\}.

\vspace{3pt}
The first commentary result: \{first\_commentary\}.

\vspace{3pt}
Based on the initial response and commentary, please further debate and refine the analysis. If there are any conflicting opinions or uncertainties, resolve them (both entity and review) and provide a more detailed and accurate response.

\vspace{3pt}
Example Output format:

\vspace{3pt}
Round 3: [{“entity": “\textless entity\textgreater", “category": “\textless category\textgreater", “sentiment": “\textless label\textgreater",“Review":“\textless review\textgreater"}]

\vspace{3pt}
Please return the final output based on the above sentence in json format.

\vspace{5pt}
\end{tcolorbox}
\caption{Self-debate prompts used in this paper for triplet extraction.}
\label{fig:self-debate-prompt-triplet}
\end{figure*}

\begin{table*}[htbp]
\setlength\tabcolsep{2pt}
\centering
\scriptsize
\begin{tabular}{ll|ccccccccc|c}
\toprule 
Round&Lang.& Claude-3.5 & Gemma-2 & Gemini-1.5 & GPT-4o & Llama-3.1 & Mistral &  Phi-3.5 & Qwen-2.5 & Zephyr& Avg.\\
\midrule
\multirow{6}{*}{1}
&EN    & 53.37& 56.52& 60.39 & 57.10& 45.60 & 48.20 & 52.19& 53.91& 39.66& 51.88\\
&ES    & 44.57& 48.71& 48.80 & 52.25& 32.49 & 38.34 & 40.08& 46.46& 33.79& 42.83\\
&FR    & 43.70& 48.74& 50.94 & 49.05& 23.15 & 37.16 & 39.81& 42.94& 30.91& 40.71\\
&NL    & 40.76& 46.45& 50.24 & 47.73& 33.53 & 33.98 & 37.41& 40.70& 26.76& 39.73\\
&RU    & 34.91& 40.38& 39.34 & 43.24& 30.18 & 26.59 & 28.71& 34.54& 23.79& 33.52\\

\cmidrule(l){2-12}
&Avg.  & 43.88  & 48.95  & 50.94 & 50.49  & 34.71 & 35.91   & 36.60& 42.38& 31.95  & 41.76\\

\midrule
\multirow{6}{*}{2} 
&EN    & 51.48& 52.82& 55.83 & 55.73& 41.76 & 42.86 & 50.06& 52.83& 41.15& 49.39\\
&ES    & 43.47& 45.59& 45.94 & 50.05& 32.07 & 36.15 & 40.02& 44.72& 34.31& 41.37\\
&FR    & 43.10& 46.23& 49.22 & 47.40& 22.88 & 33.68 & 39.83& 41.42& 32.97& 39.64\\
&NL    & 39.94& 41.15& 44.73 & 45.55& 32.04 & 30.03 & 37.71& 37.35& 27.50& 37.33\\
&RU    & 34.94& 37.44& 37.12 & 40.77& 28.80 & 23.25 & 28.31& 32.10& 23.94& 31.85\\
\cmidrule(l){2-12}
&Avg.  & 42.59& 44.65& 46.57 & 47.90& 31.51 & 33.19 & 39.19& 41.69& 31.97& 39.92\\

\midrule 
\multirow{6}{*}{3} 
&EN    & 51.69& 47.69& 53.07 & 50.18& 38.83 & 38.10 & 48.68& 49.14& 40.61& 46.44\\
&ES    & 42.69& 42.56& 43.66 & 42.83& 31.10 & 31.50 & 38.29& 40.84& 34.38& 38.65\\
&FR    & 42.05& 42.42& 45.70 & 41.09& 21.72 & 29.80 & 37.87& 38.23& 32.75& 36.85\\
&NL    & 38.52& 37.53& 42.91 & 37.17& 30.71 & 26.20 & 34.68& 34.72& 27.45& 34.43\\
&RU    & 34.26& 33.51& 35.05 & 34.69& 26.86 & 21.23 & 27.03& 29.11& 23.73& 29.50\\
\cmidrule(l){2-12}
&Avg.  & 41.84& 40.74& 44.08 & 41.19& 29.84 & 29.37 & 37.31& 38.41& 31.78& 37.17\\

\bottomrule 
\end{tabular}
\caption{Detailed performance comparison of multiple runs in self-improvement.}
\label{tab:debate}
\end{table*}

\begin{table*}[htbp]
\setlength\tabcolsep{2pt}
\centering
\scriptsize
\begin{tabular}{ll|ccccccccc|c}
\toprule 
Round&Lang.& Claude-3.5 & Gemma-2 & Gemini-1.5 & GPT-4o & Llama-3.1 & Mistral &  Phi-3.5 & Qwen-2.5 & Zephyr& Avg.\\
\midrule
\multirow{6}{*}{1}
&EN    & 54.87  & 57.35  & 61.14 & 53.72  & 47.05 & 47.81   & 48.39& 53.20& 40.45  & 51.55\\
&ES    & 43.02  & 48.60  & 49.84 & 51.71  & 35.18 & 36.42   & 39.24& 45.75& 34.36  & 42.68\\
&FR    & 43.85  & 49.44  & 50.74 & 49.18  & 25.41 & 36.48   & 36.13& 42.13& 32.64  & 40.67\\
&NL    & 42.55  & 48.16  & 52.05 & 51.80  & 34.68 & 31.82   & 33.22& 38.21& 28.04  & 40.06\\
&RU    & 35.13  & 41.19  & 40.92 & 46.07  & 31.20 & 27.03   & 26.00& 32.61& 24.28  & 33.82\\

\cmidrule(l){2-12}
&Avg.  & 43.88  & 48.95  & 50.94 & 50.49  & 34.71 & 35.91   & 36.60& 42.38& 31.95  & 41.76\\

\midrule
\multirow{6}{*}{2} 
&EN    & 52.64  & 45.19  & 49.10 & 46.40  & 34.05 & 36.47   & 47.61& 46.96& 40.66  & 44.34\\
&ES    & 40.25  & 39.91  & 40.44 & 40.03  & 24.96 & 30.48   & 37.00& 39.91& 32.68  & 36.18\\
&FR    & 39.96  & 40.43  & 41.85 & 38.65  & 17.48 & 28.68   & 34.65& 34.14& 30.97  & 34.09\\
&NL    & 39.62  & 38.55  & 40.09 & 36.76  & 25.78 & 25.33   & 31.73& 31.63& 26.77  & 32.92\\
&RU    & 32.07  & 32.55  & 34.81 & 32.87  & 22.50 & 21.54   & 24.85& 27.99& 22.25  & 27.94\\
\cmidrule(l){2-12}
&Avg.  & 40.91  & 39.33  & 41.26 & 38.94  & 24.96 & 28.50   & 35.17& 36.13& 30.67  & 35.09\\
\midrule 
\multirow{6}{*}{3} 
&EN    & 51.44  & 40.93  & 46.86 & 45.86  & 25.30 & 27.00   & 45.88& 42.59& 35.78  & 40.18\\
&ES    & 40.03  & 36.89  & 37.70 & 39.45  & 18.67 & 22.69   & 37.08& 37.05& 31.33  & 33.43\\
&FR    & 39.02  & 36.90  & 40.94 & 38.11  & 15.43 & 23.80   & 34.54& 32.02& 27.36  & 32.01\\
&NL    & 38.41  & 34.79  & 37.67 & 36.24  & 20.79 & 19.40   & 32.14& 29.81& 23.36  & 30.29\\
&RU    & 31.58  & 29.95  & 32.39 & 32.60  & 19.82 & 16.62   & 24.33& 25.43& 19.86  & 25.84\\
\cmidrule(l){2-12}
&Avg.  & 40.10  & 35.89  & 39.11 & 38.45  & 20.00 & 21.90   & 34.79& 33.38& 27.54  & 32.35\\

\bottomrule 
\end{tabular}
\caption{Detailed performance comparison of multiple runs in self-debate.}
\label{tab:improve}
\end{table*}

\begin{table*}[htbp]
\setlength\tabcolsep{2pt}
\centering
\scriptsize
\begin{tabular}{ll|ccccccccc|c}
\toprule 
Shot&Lang.& Claude-3.5 & Gemma-2 & Gemini-1.5 & GPT-4o & Llama-3.1 & Mistral &  Phi-3.5 & Qwen-2.5 & Zephyr& Avg.\\
\midrule
\multirow{6}{*}{0}
& EN    & 53.37  & 53.95  & 60.39 & 55.81  & 45.60 & 48.26   & 52.19   & 54.64& 39.15  & 51.49\\
& ES    & 44.36  & 48.29  & 48.80 & 49.91  & 32.49 & 38.32   & 40.08   & 48.10& 33.48  & 42.65\\
& FR    & 42.73  & 48.88  & 50.94 & 48.43  & 23.15 & 37.21   & 39.81   & 43.60& 32.21  & 40.77\\
& NL    & 42.03  & 46.75  & 50.24 & 49.94  & 33.53 & 33.98   & 37.41   & 42.59& 26.79  & 40.36\\
& RU    & 33.90  & 40.25  & 39.34 & 45.15  & 30.18 & 26.58   & 28.71   & 37.15& 22.67  & 33.77\\

\cmidrule(l){2-12}
& Avg.  & 43.28  & 47.63  & 49.94 & 49.85  & 32.99 & 36.87   & 39.64   & 45.22& 30.86  & 41.81\\

\midrule
\multirow{6}{*}{1} 
& EN    & 54.77  & 56.86  & 59.08 & 59.06  & 50.17 & 47.93   & 50.12   & 55.74& 39.58  & 52.59\\
& ES    & 48.33  & 50.37  & 52.16 & 57.08  & 40.07 & 41.70   & 40.98   & 48.23& 32.27  & 45.69\\
& FR    & 52.12  & 50.21  & 52.19 & 56.19  & 37.95 & 38.65   & 39.14   & 48.51& 31.39  & 45.15\\
& NL    & 43.30  & 46.49  & 48.52 & 51.26  & 34.97 & 36.42   & 35.00   & 41.64& 24.82  & 40.27\\
& RU    & 39.97  & 42.62  & 43.23 & 48.65  & 33.92 & 31.67   & 31.11   & 38.50& 24.31  & 37.11\\
\cmidrule(l){2-12}
& Avg.  & 47.70  & 49.31  & 51.04 & 54.45  & 39.42 & 39.28   & 39.27   & 46.52& 30.47  & 44.16\\

\midrule 
\multirow{6}{*}{2} 
& EN    & 57.07  & 56.93  & 59.71 & 63.88  & 51.74 & 50.20   & 52.95   & 57.90& 36.58  & 54.11\\
& ES    & 50.24  & 49.62  & 53.37 & 59.14  & 42.28 & 42.59   & 41.58   & 49.74& 29.97  & 46.50\\
& FR    & 51.84  & 50.08  & 53.53 & 56.89  & 39.34 & 40.22   & 40.74   & 50.18& 30.56  & 45.93\\
& NL    & 44.88  & 44.91  & 48.28 & 56.46  & 35.45 & 36.12   & 36.65   & 42.50& 25.58  & 41.20\\
& RU    & 41.55  & 44.48  & 43.87 & 52.66  & 35.51 & 31.90   & 30.96   & 39.48& 24.53  & 38.33\\
\cmidrule(l){2-12}
& Avg.  & 49.12  & 49.20  & 51.75 & 57.81  & 40.86 & 40.21   & 40.58   & 47.96& 29.44  & 45.21\\

\midrule 
\multirow{6}{*}{4} 
& EN    & 59.39  & 57.21  & 61.16 & 66.87  & 53.78 & 52.30   & 53.68   & 59.66& 40.76  & 56.09\\
& ES    & 52.13  & 52.69  & 55.93 & 62.57  & 44.37 & 44.03   & 43.22   & 51.91& 34.22  & 49.01\\
& FR    & 53.36  & 51.85  & 55.26 & 56.49  & 42.09 & 43.03   & 42.02   & 53.28& 32.45  & 47.76\\
& NL    & 46.49  & 46.15  & 47.97 & 56.40  & 38.65 & 37.88   & 36.10   & 42.70& 28.61  & 42.33\\
& RU    & 43.02  & 45.52  & 45.38 & 54.50  & 36.51 & 34.44   & 33.13   & 41.34& 25.65  & 39.94\\
\cmidrule(l){2-12}
& Avg.  & 50.88  & 50.69  & 53.14 & 59.37  & 43.08 & 42.33   & 41.63   & 49.78& 32.34  & 47.03\\

\midrule 
\multirow{6}{*}{8} 
& EN    & 59.34  & 59.10  & 61.15 & 68.84  & 54.30 & 51.96   & 54.24   & 61.34& 39.61  & 56.65\\
& ES    & 53.32  & 54.49  & 55.68 & 63.04  & 45.64 & 44.55   & 44.35   & 53.42& 30.19  & 49.41\\
& FR    & 55.02  & 53.33  & 56.22 & 59.02  & 43.82 & 43.44   & 42.32   & 53.50& 32.00  & 48.74\\
& NL    & 47.60  & 46.89  & 48.56 & 59.09  & 39.91 & 36.35   & 36.86   & 43.18& 25.85  & 42.70\\
& RU    & 43.43  & 46.36  & 45.94 & 55.86  & 36.79 & 35.03   & 33.43   & 42.52& 23.48  & 40.32\\
\cmidrule(l){2-12}
& Avg.  & 51.74  & 52.03  & 53.51 & 61.17  & 44.09 & 42.27   & 42.24   & 50.79& 30.23  & 47.56\\

\midrule 
\multirow{6}{*}{16} 
& EN    & 59.61  & 60.62  & 61.19 & 69.67  & 53.65 & 52.25   & 55.27   & 58.36& 36.23  & 56.32\\
& ES    & 54.06  & 55.00  & 55.11 & 64.04  & 46.07 & 46.04   & 44.99   & 54.05& 31.28  & 50.07\\
& FR    & 55.03  & 55.35  & 56.31 & 61.85  & 44.93 & 45.38   & 42.87   & 54.30& 32.24  & 49.81\\
& NL    & 46.79  & 48.22  & 48.63 & 61.13  & 39.62 & 37.07   & 38.83   & 44.35& 24.70  & 43.26\\
& RU    & 45.76  & 49.11  & 46.75 & 58.41  & 38.73 & 35.13   & 33.88   & 43.66& 24.09  & 41.73\\
\cmidrule(l){2-12}
 & Avg.  & 52.25  & 53.66  & 53.60 & 63.02  & 44.60 & 43.17   & 43.17   & 50.94& 29.71  & 48.24   \\
\bottomrule 
\end{tabular}
\caption{Detailed performance comparison of few-shot results.}
\label{tab:fewshot}
\end{table*}






\end{appendices}

\clearpage

\bibliography{sn-bibliography}


\begin{thebibliography}{56}
\ifx \bisbn   \undefined \def \bisbn  #1{ISBN #1}\fi
\ifx \binits  \undefined \def \binits#1{#1}\fi
\ifx \bauthor  \undefined \def \bauthor#1{#1}\fi
\ifx \batitle  \undefined \def \batitle#1{#1}\fi
\ifx \bjtitle  \undefined \def \bjtitle#1{#1}\fi
\ifx \bvolume  \undefined \def \bvolume#1{\textbf{#1}}\fi
\ifx \byear  \undefined \def \byear#1{#1}\fi
\ifx \bissue  \undefined \def \bissue#1{#1}\fi
\ifx \bfpage  \undefined \def \bfpage#1{#1}\fi
\ifx \blpage  \undefined \def \blpage #1{#1}\fi
\ifx \burl  \undefined \def \burl#1{\textsf{#1}}\fi
\ifx \doiurl  \undefined \def \doiurl#1{\url{https://doi.org/#1}}\fi
\ifx \betal  \undefined \def \betal{\textit{et al.}}\fi
\ifx \binstitute  \undefined \def \binstitute#1{#1}\fi
\ifx \binstitutionaled  \undefined \def \binstitutionaled#1{#1}\fi
\ifx \bctitle  \undefined \def \bctitle#1{#1}\fi
\ifx \beditor  \undefined \def \beditor#1{#1}\fi
\ifx \bpublisher  \undefined \def \bpublisher#1{#1}\fi
\ifx \bbtitle  \undefined \def \bbtitle#1{#1}\fi
\ifx \bedition  \undefined \def \bedition#1{#1}\fi
\ifx \bseriesno  \undefined \def \bseriesno#1{#1}\fi
\ifx \blocation  \undefined \def \blocation#1{#1}\fi
\ifx \bsertitle  \undefined \def \bsertitle#1{#1}\fi
\ifx \bsnm \undefined \def \bsnm#1{#1}\fi
\ifx \bsuffix \undefined \def \bsuffix#1{#1}\fi
\ifx \bparticle \undefined \def \bparticle#1{#1}\fi
\ifx \barticle \undefined \def \barticle#1{#1}\fi
\bibcommenthead
\ifx \bconfdate \undefined \def \bconfdate #1{#1}\fi
\ifx \botherref \undefined \def \botherref #1{#1}\fi
\ifx \url \undefined \def \url#1{\textsf{#1}}\fi
\ifx \bchapter \undefined \def \bchapter#1{#1}\fi
\ifx \bbook \undefined \def \bbook#1{#1}\fi
\ifx \bcomment \undefined \def \bcomment#1{#1}\fi
\ifx \oauthor \undefined \def \oauthor#1{#1}\fi
\ifx \citeauthoryear \undefined \def \citeauthoryear#1{#1}\fi
\ifx \endbibitem  \undefined \def \endbibitem {}\fi
\ifx \bconflocation  \undefined \def \bconflocation#1{#1}\fi
\ifx \arxivurl  \undefined \def \arxivurl#1{\textsf{#1}}\fi
\csname PreBibitemsHook\endcsname

\bibitem[\protect\citeauthoryear{Liu}{2012}]{liu2012}
\begin{bbook}
\bauthor{\bsnm{Liu}, \binits{B.}}:
\bbtitle{Sentiment Analysis and Opinion Mining}.
\bsertitle{Synthesis Lectures on Human Language Technologies},
(\byear{2012}).
\doiurl{10.2200/S00416ED1V01Y201204HLT016} .
\burl{https://doi.org/10.2200/S00416ED1V01Y201204HLT016}
\end{bbook}
\endbibitem

\bibitem[\protect\citeauthoryear{Ng et~al.}{}]{ngcharacterising}
\begin{botherref}
\oauthor{\bsnm{Ng}, \binits{S.}},
\oauthor{\bsnm{Rahman}, \binits{N.}},
\oauthor{\bsnm{Ang}, \binits{I.}},
\oauthor{\bsnm{Sridharan}, \binits{S.}},
\oauthor{\bsnm{Ramachandran}, \binits{S.}},
\oauthor{\bsnm{Wang}, \binits{D.}},
\oauthor{\bsnm{Khoo}, \binits{A.}},
\oauthor{\bsnm{Tan}, \binits{C.}},
\oauthor{\bsnm{Feng}, \binits{M.}},
\oauthor{\bsnm{Toh}, \binits{S.}}, et al.:
Characterising and predicting persistent high-cost utilisers in healthcare: a retrospective cohort study in Singapore. BMJ Open. 2020 Jan 06; 10 (1): e031622. 10.1136/bmjopen-2019-031622
\end{botherref}
\endbibitem

\bibitem[\protect\citeauthoryear{Zhu et~al.}{2021}]{zhu2021genotype}
\begin{barticle}
\bauthor{\bsnm{Zhu}, \binits{M.}},
\bauthor{\bsnm{Wang}, \binits{D.D.}},
\bauthor{\bsnm{Yan}, \binits{H.}}:
\batitle{Genotype-determined egfr-rtk heterodimerization and its effects on drug resistance in lung cancer treatment revealed by molecular dynamics simulations}.
\bjtitle{BMC molecular and cell biology}
\bvolume{22}(\bissue{1}),
\bfpage{34}
(\byear{2021})
\end{barticle}
\endbibitem

\bibitem[\protect\citeauthoryear{Rahman et~al.}{2018}]{rahman2018processing}
\begin{barticle}
\bauthor{\bsnm{Rahman}, \binits{N.}},
\bauthor{\bsnm{Wang}, \binits{D.D.}},
\bauthor{\bsnm{Ng}, \binits{S.H.-X.}},
\bauthor{\bsnm{Ramachandran}, \binits{S.}},
\bauthor{\bsnm{Sridharan}, \binits{S.}},
\bauthor{\bsnm{Khoo}, \binits{A.}},
\bauthor{\bsnm{Tan}, \binits{C.S.}},
\bauthor{\bsnm{Goh}, \binits{W.-P.}},
\bauthor{\bsnm{Tan}, \binits{X.Q.}}, \betal:
\batitle{Processing of electronic medical records for health services research in an academic medical center: methods and validation}.
\bjtitle{JMIR Medical Informatics}
\bvolume{6}(\bissue{4}),
\bfpage{10933}
(\byear{2018})
\end{barticle}
\endbibitem

\bibitem[\protect\citeauthoryear{Mao et~al.}{2022}]{mao2022}
\begin{botherref}
\oauthor{\bsnm{Mao}, \binits{Y.}},
\oauthor{\bsnm{Shen}, \binits{Y.}},
\oauthor{\bsnm{Yu}, \binits{C.}},
\oauthor{\bsnm{Cai}, \binits{L.}}:
A joint training dual-mrc framework for aspect based sentiment analysis.
Proceedings of the AAAI Conference on Artificial Intelligence,
13543--13551
(2022)
\doiurl{10.1609/aaai.v35i15.17597}
\end{botherref}
\endbibitem

\bibitem[\protect\citeauthoryear{Zhang et~al.}{2021}]{zhang2021towards}
\begin{bchapter}
\bauthor{\bsnm{Zhang}, \binits{W.}},
\bauthor{\bsnm{Li}, \binits{X.}},
\bauthor{\bsnm{Deng}, \binits{Y.}},
\bauthor{\bsnm{Bing}, \binits{L.}},
\bauthor{\bsnm{Lam}, \binits{W.}}:
\bctitle{Towards generative aspect-based sentiment analysis}.
In: \beditor{\bsnm{Zong}, \binits{C.}},
\beditor{\bsnm{Xia}, \binits{F.}},
\beditor{\bsnm{Li}, \binits{W.}},
\beditor{\bsnm{Navigli}, \binits{R.}} (eds.)
\bbtitle{Proceedings of the 59th Annual Meeting of the Association for Computational Linguistics and the 11th International Joint Conference on Natural Language Processing (Volume 2: Short Papers)},
pp. \bfpage{504}--\blpage{510}.
\bpublisher{Association for Computational Linguistics},
\blocation{Online}
(\byear{2021}).
\doiurl{10.18653/v1/2021.acl-short.64} .
\burl{https://aclanthology.org/2021.acl-short.64}
\end{bchapter}
\endbibitem

\bibitem[\protect\citeauthoryear{Tran and Matsui}{2024}]{tran2024improving}
\begin{bchapter}
\bauthor{\bsnm{Tran}, \binits{V.}},
\bauthor{\bsnm{Matsui}, \binits{T.}}:
\bctitle{Improving llm prompting with ensemble of instructions: A case study on sentiment analysis}.
In: \bbtitle{JSAI International Symposium on Artificial Intelligence},
pp. \bfpage{299}--\blpage{305}
(\byear{2024}).
\bcomment{Springer}
\end{bchapter}
\endbibitem

\bibitem[\protect\citeauthoryear{van Schaik and Pugh}{2024}]{van2024field}
\begin{bchapter}
\bauthor{\bsnm{Schaik}, \binits{T.A.}},
\bauthor{\bsnm{Pugh}, \binits{B.}}:
\bctitle{A field guide to automatic evaluation of llm-generated summaries}.
In: \bbtitle{Proceedings of the 47th International ACM SIGIR Conference on Research and Development in Information Retrieval},
pp. \bfpage{2832}--\blpage{2836}
(\byear{2024})
\end{bchapter}
\endbibitem

\bibitem[\protect\citeauthoryear{Roumeliotis et~al.}{2024}]{roumeliotis2024llms}
\begin{barticle}
\bauthor{\bsnm{Roumeliotis}, \binits{K.I.}},
\bauthor{\bsnm{Tselikas}, \binits{N.D.}},
\bauthor{\bsnm{Nasiopoulos}, \binits{D.K.}}:
\batitle{Llms in e-commerce: a comparative analysis of gpt and llama models in product review evaluation}.
\bjtitle{Natural Language Processing Journal}
\bvolume{6},
\bfpage{100056}
(\byear{2024})
\end{barticle}
\endbibitem

\bibitem[\protect\citeauthoryear{Xie et~al.}{2023}]{xie-etal-2023-empirical}
\begin{bchapter}
\bauthor{\bsnm{Xie}, \binits{T.}},
\bauthor{\bsnm{Li}, \binits{Q.}},
\bauthor{\bsnm{Zhang}, \binits{J.}},
\bauthor{\bsnm{Zhang}, \binits{Y.}},
\bauthor{\bsnm{Liu}, \binits{Z.}},
\bauthor{\bsnm{Wang}, \binits{H.}}:
\bctitle{Empirical study of zero-shot {NER} with {C}hat{GPT}}.
In: \beditor{\bsnm{Bouamor}, \binits{H.}},
\beditor{\bsnm{Pino}, \binits{J.}},
\beditor{\bsnm{Bali}, \binits{K.}} (eds.)
\bbtitle{Proceedings of the 2023 Conference on Empirical Methods in Natural Language Processing},
pp. \bfpage{7935}--\blpage{7956}.
\bpublisher{Association for Computational Linguistics},
\blocation{Singapore}
(\byear{2023}).
\doiurl{10.18653/v1/2023.emnlp-main.493} .
\burl{https://aclanthology.org/2023.emnlp-main.493}
\end{bchapter}
\endbibitem

\bibitem[\protect\citeauthoryear{Ma et~al.}{2024}]{ma-etal-2024-topro}
\begin{bchapter}
\bauthor{\bsnm{Ma}, \binits{B.}},
\bauthor{\bsnm{Nie}, \binits{E.}},
\bauthor{\bsnm{Yuan}, \binits{S.}},
\bauthor{\bsnm{Schmid}, \binits{H.}},
\bauthor{\bsnm{F{\"a}rber}, \binits{M.}},
\bauthor{\bsnm{Kreuter}, \binits{F.}},
\bauthor{\bsnm{Schuetze}, \binits{H.}}:
\bctitle{{T}o{P}ro: Token-level prompt decomposition for cross-lingual sequence labeling tasks}.
In: \beditor{\bsnm{Graham}, \binits{Y.}},
\beditor{\bsnm{Purver}, \binits{M.}} (eds.)
\bbtitle{Proceedings of the 18th Conference of the European Chapter of the Association for Computational Linguistics (Volume 1: Long Papers)},
pp. \bfpage{2685}--\blpage{2702}.
\bpublisher{Association for Computational Linguistics},
\blocation{St. Julian{'}s, Malta}
(\byear{2024}).
\burl{https://aclanthology.org/2024.eacl-long.164}
\end{bchapter}
\endbibitem

\bibitem[\protect\citeauthoryear{Nie et~al.}{2024}]{nie2024decomposed}
\begin{botherref}
\oauthor{\bsnm{Nie}, \binits{E.}},
\oauthor{\bsnm{Yuan}, \binits{S.}},
\oauthor{\bsnm{Ma}, \binits{B.}},
\oauthor{\bsnm{Schmid}, \binits{H.}},
\oauthor{\bsnm{Färber}, \binits{M.}},
\oauthor{\bsnm{Kreuter}, \binits{F.}},
\oauthor{\bsnm{Schütze}, \binits{H.}}:
Decomposed Prompting: Unveiling Multilingual Linguistic Structure Knowledge in English-Centric Large Language Models
(2024).
\url{https://arxiv.org/abs/2402.18397}
\end{botherref}
\endbibitem

\bibitem[\protect\citeauthoryear{Zhang et~al.}{2024}]{zhang-etal-2024-sentiment}
\begin{bchapter}
\bauthor{\bsnm{Zhang}, \binits{W.}},
\bauthor{\bsnm{Deng}, \binits{Y.}},
\bauthor{\bsnm{Liu}, \binits{B.}},
\bauthor{\bsnm{Pan}, \binits{S.}},
\bauthor{\bsnm{Bing}, \binits{L.}}:
\bctitle{Sentiment analysis in the era of large language models: A reality check}.
In: \beditor{\bsnm{Duh}, \binits{K.}},
\beditor{\bsnm{Gomez}, \binits{H.}},
\beditor{\bsnm{Bethard}, \binits{S.}} (eds.)
\bbtitle{Findings of the Association for Computational Linguistics: NAACL 2024},
pp. \bfpage{3881}--\blpage{3906}.
\bpublisher{Association for Computational Linguistics},
\blocation{Mexico City, Mexico}
(\byear{2024}).
\doiurl{10.18653/v1/2024.findings-naacl.246} .
\burl{https://aclanthology.org/2024.findings-naacl.246}
\end{bchapter}
\endbibitem

\bibitem[\protect\citeauthoryear{Pontiki et~al.}{2016}]{pontiki-etal-2016-semeval}
\begin{bchapter}
\bauthor{\bsnm{Pontiki}, \binits{M.}},
\bauthor{\bsnm{Galanis}, \binits{D.}},
\bauthor{\bsnm{Papageorgiou}, \binits{H.}},
\bauthor{\bsnm{Androutsopoulos}, \binits{I.}},
\bauthor{\bsnm{Manandhar}, \binits{S.}},
\bauthor{\bsnm{AL-Smadi}, \binits{M.}},
\bauthor{\bsnm{Al-Ayyoub}, \binits{M.}},
\bauthor{\bsnm{Zhao}, \binits{Y.}},
\bauthor{\bsnm{Qin}, \binits{B.}},
\bauthor{\bsnm{De~Clercq}, \binits{O.}},
\bauthor{\bsnm{Hoste}, \binits{V.}},
\bauthor{\bsnm{Apidianaki}, \binits{M.}},
\bauthor{\bsnm{Tannier}, \binits{X.}},
\bauthor{\bsnm{Loukachevitch}, \binits{N.}},
\bauthor{\bsnm{Kotelnikov}, \binits{E.}},
\bauthor{\bsnm{Bel}, \binits{N.}},
\bauthor{\bsnm{Jim{\'e}nez-Zafra}, \binits{S.M.}},
\bauthor{\bsnm{Eryi{\u{g}}it}, \binits{G.}}:
\bctitle{{S}em{E}val-2016 task 5: Aspect based sentiment analysis}.
In: \beditor{\bsnm{Bethard}, \binits{S.}},
\beditor{\bsnm{Carpuat}, \binits{M.}},
\beditor{\bsnm{Cer}, \binits{D.}},
\beditor{\bsnm{Jurgens}, \binits{D.}},
\beditor{\bsnm{Nakov}, \binits{P.}},
\beditor{\bsnm{Zesch}, \binits{T.}} (eds.)
\bbtitle{Proceedings of the 10th International Workshop on Semantic Evaluation ({S}em{E}val-2016)},
pp. \bfpage{19}--\blpage{30}.
\bpublisher{Association for Computational Linguistics},
\blocation{San Diego, California}
(\byear{2016}).
\doiurl{10.18653/v1/S16-1002} .
\burl{https://aclanthology.org/S16-1002}
\end{bchapter}
\endbibitem

\bibitem[\protect\citeauthoryear{Lin et~al.}{2014}]{lin2014cross}
\begin{bchapter}
\bauthor{\bsnm{Lin}, \binits{Z.}},
\bauthor{\bsnm{Jin}, \binits{X.}},
\bauthor{\bsnm{Xu}, \binits{X.}},
\bauthor{\bsnm{Wang}, \binits{W.}},
\bauthor{\bsnm{Cheng}, \binits{X.}},
\bauthor{\bsnm{Wang}, \binits{Y.}}:
\bctitle{A cross-lingual joint aspect/sentiment model for sentiment analysis}.
In: \bbtitle{Proceedings of the 23rd ACM International Conference on Conference on Information and Knowledge Management},
pp. \bfpage{1089}--\blpage{1098}
(\byear{2014})
\end{bchapter}
\endbibitem

\bibitem[\protect\citeauthoryear{Klinger and Cimiano}{2015}]{klinger2015instance}
\begin{bchapter}
\bauthor{\bsnm{Klinger}, \binits{R.}},
\bauthor{\bsnm{Cimiano}, \binits{P.}}:
\bctitle{Instance selection improves cross-lingual model training for fine-grained sentiment analysis}.
In: \bbtitle{Proceedings of the Nineteenth Conference on Computational Natural Language Learning},
pp. \bfpage{153}--\blpage{163}.
\bpublisher{Association for Computational Linguistics},
\blocation{Beijing, China}
(\byear{2015}).
\doiurl{10.18653/v1/K15-1016} .
\burl{https://aclanthology.org/K15-1016}
\end{bchapter}
\endbibitem

\bibitem[\protect\citeauthoryear{Lambert}{2015}]{lambert2015aspect}
\begin{bchapter}
\bauthor{\bsnm{Lambert}, \binits{P.}}:
\bctitle{Aspect-level cross-lingual sentiment classification with constrained {SMT}}.
In: \beditor{\bsnm{Zong}, \binits{C.}},
\beditor{\bsnm{Strube}, \binits{M.}} (eds.)
\bbtitle{Proceedings of the 53rd Annual Meeting of the Association for Computational Linguistics and the 7th International Joint Conference on Natural Language Processing (Volume 2: Short Papers)},
pp. \bfpage{781}--\blpage{787}.
\bpublisher{Association for Computational Linguistics},
\blocation{Beijing, China}
(\byear{2015}).
\doiurl{10.3115/v1/P15-2128} .
\burl{https://aclanthology.org/P15-2128}
\end{bchapter}
\endbibitem

\bibitem[\protect\citeauthoryear{Barnes et~al.}{2016}]{barnes2016exploring}
\begin{bchapter}
\bauthor{\bsnm{Barnes}, \binits{J.}},
\bauthor{\bsnm{Lambert}, \binits{P.}},
\bauthor{\bsnm{Badia}, \binits{T.}}:
\bctitle{Exploring distributional representations and machine translation for aspect-based cross-lingual sentiment classification.}
In: \beditor{\bsnm{Matsumoto}, \binits{Y.}},
\beditor{\bsnm{Prasad}, \binits{R.}} (eds.)
\bbtitle{Proceedings of {COLING} 2016, the 26th International Conference on Computational Linguistics: Technical Papers},
pp. \bfpage{1613}--\blpage{1623}.
\bpublisher{The COLING 2016 Organizing Committee},
\blocation{Osaka, Japan}
(\byear{2016}).
\burl{https://aclanthology.org/C16-1152}
\end{bchapter}
\endbibitem

\bibitem[\protect\citeauthoryear{Phan et~al.}{2021}]{9585242}
\begin{bchapter}
\bauthor{\bsnm{Phan}, \binits{K.T.-K.}},
\bauthor{\bsnm{Ngoc~Hao}, \binits{D.}},
\bauthor{\bsnm{Thin}, \binits{D.V.}},
\bauthor{\bsnm{Luu-Thuy~Nguyen}, \binits{N.}}:
\bctitle{Exploring zero-shot cross-lingual aspect-based sentiment analysis using pre-trained multilingual language models}.
In: \bbtitle{2021 International Conference on Multimedia Analysis and Pattern Recognition (MAPR)},
pp. \bfpage{1}--\blpage{6}
(\byear{2021}).
\doiurl{10.1109/MAPR53640.2021.9585242}
\end{bchapter}
\endbibitem

\bibitem[\protect\citeauthoryear{Xu et~al.}{2019}]{xu2019bert}
\begin{bchapter}
\bauthor{\bsnm{Xu}, \binits{H.}},
\bauthor{\bsnm{Liu}, \binits{B.}},
\bauthor{\bsnm{Shu}, \binits{L.}},
\bauthor{\bsnm{Yu}, \binits{P.}}:
\bctitle{{BERT} post-training for review reading comprehension and aspect-based sentiment analysis}.
In: \beditor{\bsnm{Burstein}, \binits{J.}},
\beditor{\bsnm{Doran}, \binits{C.}},
\beditor{\bsnm{Solorio}, \binits{T.}} (eds.)
\bbtitle{Proceedings of the 2019 Conference of the North {A}merican Chapter of the Association for Computational Linguistics: Human Language Technologies, Volume 1 (Long and Short Papers)},
pp. \bfpage{2324}--\blpage{2335}.
\bpublisher{Association for Computational Linguistics},
\blocation{Minneapolis, Minnesota}
(\byear{2019}).
\doiurl{10.18653/v1/N19-1242} .
\burl{https://aclanthology.org/N19-1242}
\end{bchapter}
\endbibitem

\bibitem[\protect\citeauthoryear{Zhao and Yu}{2021}]{zhao2021knowledge}
\begin{barticle}
\bauthor{\bsnm{Zhao}, \binits{A.}},
\bauthor{\bsnm{Yu}, \binits{Y.}}:
\batitle{Knowledge-enabled bert for aspect-based sentiment analysis}.
\bjtitle{Knowledge-Based Systems}
\bvolume{227},
\bfpage{107220}
(\byear{2021})
\doiurl{10.1016/j.knosys.2021.107220}
\end{barticle}
\endbibitem

\bibitem[\protect\citeauthoryear{Zhang et~al.}{2021}]{zhang2021cross}
\begin{bchapter}
\bauthor{\bsnm{Zhang}, \binits{W.}},
\bauthor{\bsnm{He}, \binits{R.}},
\bauthor{\bsnm{Peng}, \binits{H.}},
\bauthor{\bsnm{Bing}, \binits{L.}},
\bauthor{\bsnm{Lam}, \binits{W.}}:
\bctitle{Cross-lingual aspect-based sentiment analysis with aspect term code-switching}.
In: \beditor{\bsnm{Moens}, \binits{M.}},
\beditor{\bsnm{Huang}, \binits{X.}},
\beditor{\bsnm{Specia}, \binits{L.}},
\beditor{\bsnm{Yih}, \binits{S.W.}} (eds.)
\bbtitle{Proceedings of the 2021 Conference on Empirical Methods in Natural Language Processing, {EMNLP} 2021, Virtual Event / Punta Cana, Dominican Republic, 7-11 November, 2021},
pp. \bfpage{9220}--\blpage{9230}.
\bpublisher{Association for Computational Linguistics}, \blocation{???}
(\byear{2021}).
\doiurl{10.18653/V1/2021.EMNLP-MAIN.727} .
\burl{https://doi.org/10.18653/v1/2021.emnlp-main.727}
\end{bchapter}
\endbibitem

\bibitem[\protect\citeauthoryear{Lin et~al.}{2023}]{lin2023}
\begin{barticle}
\bauthor{\bsnm{Lin}, \binits{N.}},
\bauthor{\bsnm{Fu}, \binits{Y.}},
\bauthor{\bsnm{Lin}, \binits{X.}},
\bauthor{\bsnm{Zhou}, \binits{D.}},
\bauthor{\bsnm{Yang}, \binits{A.}},
\bauthor{\bsnm{Jiang}, \binits{S.}}:
\batitle{Cl-xabsa: Contrastive learning for cross-lingual aspect-based sentiment analysis}.
\bjtitle{IEEE/ACM Transactions on Audio, Speech, and Language Processing}
\bvolume{31},
\bfpage{2935}--\blpage{2946}
(\byear{2023})
\doiurl{10.1109/TASLP.2023.3297964}
\end{barticle}
\endbibitem

\bibitem[\protect\citeauthoryear{Zhang et~al.}{2021}]{zhang-etal-2021-cross}
\begin{bchapter}
\bauthor{\bsnm{Zhang}, \binits{W.}},
\bauthor{\bsnm{He}, \binits{R.}},
\bauthor{\bsnm{Peng}, \binits{H.}},
\bauthor{\bsnm{Bing}, \binits{L.}},
\bauthor{\bsnm{Lam}, \binits{W.}}:
\bctitle{Cross-lingual aspect-based sentiment analysis with aspect term code-switching}.
In: \beditor{\bsnm{Moens}, \binits{M.-F.}},
\beditor{\bsnm{Huang}, \binits{X.}},
\beditor{\bsnm{Specia}, \binits{L.}},
\beditor{\bsnm{Yih}, \binits{S.W.-t.}} (eds.)
\bbtitle{Proceedings of the 2021 Conference on Empirical Methods in Natural Language Processing},
pp. \bfpage{9220}--\blpage{9230}.
\bpublisher{Association for Computational Linguistics},
\blocation{Online and Punta Cana, Dominican Republic}
(\byear{2021}).
\doiurl{10.18653/v1/2021.emnlp-main.727} .
\burl{https://aclanthology.org/2021.emnlp-main.727}
\end{bchapter}
\endbibitem

\bibitem[\protect\citeauthoryear{Jun et~al.}{2023}]{jun2023rating}
\begin{barticle}
\bauthor{\bsnm{Jun}, \binits{Z.}},
\bauthor{\bsnm{Longlong}, \binits{Q.}},
\bauthor{\bsnm{Fanfan}, \binits{S.}},
\bauthor{\bsnm{Yueshun}, \binits{H.}},
\bauthor{\bsnm{Hai}, \binits{T.}},
\bauthor{\bsnm{Yanxiang}, \binits{H.}}:
\batitle{Rating text classification with weighted negative supervision on classifier layer}.
\bjtitle{Chinese Journal of Electronics}
\bvolume{32}(\bissue{6}),
\bfpage{1304}--\blpage{1318}
(\byear{2023})
\end{barticle}
\endbibitem

\bibitem[\protect\citeauthoryear{Zhang et~al.}{2023}]{zhang2023model}
\begin{barticle}
\bauthor{\bsnm{Zhang}, \binits{Y.}},
\bauthor{\bsnm{Li}, \binits{J.}},
\bauthor{\bsnm{Xin}, \binits{Y.}},
\bauthor{\bsnm{Zhao}, \binits{X.}},
\bauthor{\bsnm{Liu}, \binits{Y.}}:
\batitle{A model for chinese named entity recognition based on global pointer and adversarial learning}.
\bjtitle{Chinese Journal of Electronics}
\bvolume{32}(\bissue{4}),
\bfpage{854}--\blpage{867}
(\byear{2023})
\end{barticle}
\endbibitem

\bibitem[\protect\citeauthoryear{Wang et~al.}{2023}]{wang2023gpt}
\begin{botherref}
\oauthor{\bsnm{Wang}, \binits{S.}},
\oauthor{\bsnm{Sun}, \binits{X.}},
\oauthor{\bsnm{Li}, \binits{X.}},
\oauthor{\bsnm{Ouyang}, \binits{R.}},
\oauthor{\bsnm{Wu}, \binits{F.}},
\oauthor{\bsnm{Zhang}, \binits{T.}},
\oauthor{\bsnm{Li}, \binits{J.}},
\oauthor{\bsnm{Wang}, \binits{G.}}:
Gpt-ner: Named entity recognition via large language models.
arXiv preprint arXiv:2304.10428
(2023)
\end{botherref}
\endbibitem

\bibitem[\protect\citeauthoryear{Simmering and Huoviala}{2023}]{simmering2023large}
\begin{botherref}
\oauthor{\bsnm{Simmering}, \binits{P.F.}},
\oauthor{\bsnm{Huoviala}, \binits{P.}}:
Large language models for aspect-based sentiment analysis
(2023).
\url{https://arxiv.org/abs/2310.18025}
\end{botherref}
\endbibitem

\bibitem[\protect\citeauthoryear{Wang et~al.}{2023}]{wang2023selfconsistency}
\begin{botherref}
\oauthor{\bsnm{Wang}, \binits{X.}},
\oauthor{\bsnm{Wei}, \binits{J.}},
\oauthor{\bsnm{Schuurmans}, \binits{D.}},
\oauthor{\bsnm{Le}, \binits{Q.V.}},
\oauthor{\bsnm{Chi}, \binits{E.H.}},
\oauthor{\bsnm{Narang}, \binits{S.}},
\oauthor{\bsnm{Chowdhery}, \binits{A.}},
\oauthor{\bsnm{Zhou}, \binits{D.}}:
Self-Consistency Improves Chain of Thought Reasoning in Language Models
(2023).
\url{https://openreview.net/forum?id=1PL1NIMMrw}
\end{botherref}
\endbibitem

\bibitem[\protect\citeauthoryear{Wang et~al.}{2024}]{wang-etal-2024-context}
\begin{bchapter}
\bauthor{\bsnm{Wang}, \binits{Q.}},
\bauthor{\bsnm{Xu}, \binits{H.}},
\bauthor{\bsnm{Ding}, \binits{K.}},
\bauthor{\bsnm{Liang}, \binits{B.}},
\bauthor{\bsnm{Xu}, \binits{R.}}:
\bctitle{In-context example retrieval from multi-perspectives for few-shot aspect-based sentiment analysis}.
In: \beditor{\bsnm{Calzolari}, \binits{N.}},
\beditor{\bsnm{Kan}, \binits{M.-Y.}},
\beditor{\bsnm{Hoste}, \binits{V.}},
\beditor{\bsnm{Lenci}, \binits{A.}},
\beditor{\bsnm{Sakti}, \binits{S.}},
\beditor{\bsnm{Xue}, \binits{N.}} (eds.)
\bbtitle{Proceedings of the 2024 Joint International Conference on Computational Linguistics, Language Resources and Evaluation (LREC-COLING 2024)},
pp. \bfpage{8975}--\blpage{8985}.
\bpublisher{ELRA and ICCL},
\blocation{Torino, Italia}
(\byear{2024}).
\burl{https://aclanthology.org/2024.lrec-main.786}
\end{bchapter}
\endbibitem

\bibitem[\protect\citeauthoryear{Zhong et~al.}{}]{Zhong_Ding_Liu_Du_Tao}
\begin{botherref}
\oauthor{\bsnm{Zhong}, \binits{Q.}},
\oauthor{\bsnm{Ding}, \binits{L.}},
\oauthor{\bsnm{Liu}, \binits{J.}},
\oauthor{\bsnm{Du}, \binits{B.}},
\oauthor{\bsnm{Tao}, \binits{D.}}:
Can chatgpt understand too? a comparative study on chatgpt and fine-tuned bert
\end{botherref}
\endbibitem

\bibitem[\protect\citeauthoryear{Wang et~al.}{2023}]{Wang_Xie_Ding_Feng_Xia_2023}
\begin{botherref}
\oauthor{\bsnm{Wang}, \binits{Z.}},
\oauthor{\bsnm{Xie}, \binits{Q.}},
\oauthor{\bsnm{Ding}, \binits{Z.}},
\oauthor{\bsnm{Feng}, \binits{Y.}},
\oauthor{\bsnm{Xia}, \binits{R.}}:
Is chatgpt a good sentiment analyzer? a preliminary study
(2023)
\end{botherref}
\endbibitem

\bibitem[\protect\citeauthoryear{Han et~al.}{}]{Han_Peng_Yang_Wang_Liu_Wan}
\begin{botherref}
\oauthor{\bsnm{Han}, \binits{R.}},
\oauthor{\bsnm{Peng}, \binits{T.}},
\oauthor{\bsnm{Yang}, \binits{C.}},
\oauthor{\bsnm{Wang}, \binits{B.}},
\oauthor{\bsnm{Liu}, \binits{L.}},
\oauthor{\bsnm{Wan}, \binits{X.}}:
Is information extraction solved by chatgpt? an analysis of performance, evaluation criteria, robustness and errors
\end{botherref}
\endbibitem

\bibitem[\protect\citeauthoryear{Deng et~al.}{}]{Deng_Bashlovkina_Han_Baumgartner_Bendersky}
\begin{botherref}
\oauthor{\bsnm{Deng}, \binits{X.}},
\oauthor{\bsnm{Bashlovkina}, \binits{V.}},
\oauthor{\bsnm{Han}, \binits{F.}},
\oauthor{\bsnm{Baumgartner}, \binits{S.}},
\oauthor{\bsnm{Bendersky}, \binits{M.}}:
Llms to the moon? reddit market sentiment analysis with large language models
\end{botherref}
\endbibitem

\bibitem[\protect\citeauthoryear{Perez et~al.}{2021}]{perez2021}
\begin{bchapter}
\bauthor{\bsnm{Perez}, \binits{E.}},
\bauthor{\bsnm{Kiela}, \binits{D.}},
\bauthor{\bsnm{Cho}, \binits{K.}}:
\bctitle{True few-shot learning with language models}.
In: \beditor{\bsnm{Ranzato}, \binits{M.}},
\beditor{\bsnm{Beygelzimer}, \binits{A.}},
\beditor{\bsnm{Dauphin}, \binits{Y.}},
\beditor{\bsnm{Liang}, \binits{P.S.}},
\beditor{\bsnm{Vaughan}, \binits{J.W.}} (eds.)
\bbtitle{Advances in Neural Information Processing Systems},
vol. \bseriesno{34},
pp. \bfpage{11054}--\blpage{11070}.
\bpublisher{Curran Associates, Inc.}, \blocation{???}
(\byear{2021}).
\burl{https://openreview.net/forum?id=ShnM-rRh4T}
\end{bchapter}
\endbibitem

\bibitem[\protect\citeauthoryear{Lu et~al.}{2022}]{lu-etal-2022-fantastically}
\begin{bchapter}
\bauthor{\bsnm{Lu}, \binits{Y.}},
\bauthor{\bsnm{Bartolo}, \binits{M.}},
\bauthor{\bsnm{Moore}, \binits{A.}},
\bauthor{\bsnm{Riedel}, \binits{S.}},
\bauthor{\bsnm{Stenetorp}, \binits{P.}}:
\bctitle{Fantastically ordered prompts and where to find them: Overcoming few-shot prompt order sensitivity}.
In: \beditor{\bsnm{Muresan}, \binits{S.}},
\beditor{\bsnm{Nakov}, \binits{P.}},
\beditor{\bsnm{Villavicencio}, \binits{A.}} (eds.)
\bbtitle{Proceedings of the 60th Annual Meeting of the Association for Computational Linguistics (Volume 1: Long Papers)},
pp. \bfpage{8086}--\blpage{8098}.
\bpublisher{Association for Computational Linguistics},
\blocation{Dublin, Ireland}
(\byear{2022}).
\doiurl{10.18653/v1/2022.acl-long.556} .
\burl{https://aclanthology.org/2022.acl-long.556}
\end{bchapter}
\endbibitem

\bibitem[\protect\citeauthoryear{Wei et~al.}{2022}]{wei2022}
\begin{bchapter}
\bauthor{\bsnm{Wei}, \binits{J.}},
\bauthor{\bsnm{Wang}, \binits{X.}},
\bauthor{\bsnm{Schuurmans}, \binits{D.}},
\bauthor{\bsnm{Bosma}, \binits{M.}},
\bauthor{\bsnm{Ichter}, \binits{B.}},
\bauthor{\bsnm{Xia}, \binits{F.}},
\bauthor{\bsnm{Chi}, \binits{E.H.}},
\bauthor{\bsnm{Le}, \binits{Q.V.}},
\bauthor{\bsnm{Zhou}, \binits{D.}}:
\bctitle{Chain-of-thought prompting elicits reasoning in large language models}.
In: \bbtitle{Proceedings of the 36th International Conference on Neural Information Processing Systems}.
\bsertitle{NIPS '22}.
\bpublisher{Curran Associates Inc.},
\blocation{Red Hook, NY, USA}
(\byear{2022})
\end{bchapter}
\endbibitem

\bibitem[\protect\citeauthoryear{Wang et~al.}{2023}]{wang-etal-2023-towards}
\begin{bchapter}
\bauthor{\bsnm{Wang}, \binits{B.}},
\bauthor{\bsnm{Min}, \binits{S.}},
\bauthor{\bsnm{Deng}, \binits{X.}},
\bauthor{\bsnm{Shen}, \binits{J.}},
\bauthor{\bsnm{Wu}, \binits{Y.}},
\bauthor{\bsnm{Zettlemoyer}, \binits{L.}},
\bauthor{\bsnm{Sun}, \binits{H.}}:
\bctitle{Towards understanding chain-of-thought prompting: An empirical study of what matters}.
In: \beditor{\bsnm{Rogers}, \binits{A.}},
\beditor{\bsnm{Boyd-Graber}, \binits{J.}},
\beditor{\bsnm{Okazaki}, \binits{N.}} (eds.)
\bbtitle{Proceedings of the 61st Annual Meeting of the Association for Computational Linguistics (Volume 1: Long Papers)},
pp. \bfpage{2717}--\blpage{2739}.
\bpublisher{Association for Computational Linguistics},
\blocation{Toronto, Canada}
(\byear{2023}).
\doiurl{10.18653/v1/2023.acl-long.153} .
\burl{https://aclanthology.org/2023.acl-long.153}
\end{bchapter}
\endbibitem

\bibitem[\protect\citeauthoryear{Xie et~al.}{2024}]{xie-etal-2024-self}
\begin{bchapter}
\bauthor{\bsnm{Xie}, \binits{T.}},
\bauthor{\bsnm{Li}, \binits{Q.}},
\bauthor{\bsnm{Zhang}, \binits{Y.}},
\bauthor{\bsnm{Liu}, \binits{Z.}},
\bauthor{\bsnm{Wang}, \binits{H.}}:
\bctitle{Self-improving for zero-shot named entity recognition with large language models}.
In: \beditor{\bsnm{Duh}, \binits{K.}},
\beditor{\bsnm{Gomez}, \binits{H.}},
\beditor{\bsnm{Bethard}, \binits{S.}} (eds.)
\bbtitle{Proceedings of the 2024 Conference of the North American Chapter of the Association for Computational Linguistics: Human Language Technologies (Volume 2: Short Papers)},
pp. \bfpage{583}--\blpage{593}.
\bpublisher{Association for Computational Linguistics},
\blocation{Mexico City, Mexico}
(\byear{2024}).
\doiurl{10.18653/v1/2024.naacl-short.49} .
\burl{https://aclanthology.org/2024.naacl-short.49}
\end{bchapter}
\endbibitem

\bibitem[\protect\citeauthoryear{Wang et~al.}{2024}]{wang-etal-2024-rethinking-bounds}
\begin{bchapter}
\bauthor{\bsnm{Wang}, \binits{Q.}},
\bauthor{\bsnm{Wang}, \binits{Z.}},
\bauthor{\bsnm{Su}, \binits{Y.}},
\bauthor{\bsnm{Tong}, \binits{H.}},
\bauthor{\bsnm{Song}, \binits{Y.}}:
\bctitle{Rethinking the bounds of {LLM} reasoning: Are multi-agent discussions the key?}
In: \beditor{\bsnm{Ku}, \binits{L.-W.}},
\beditor{\bsnm{Martins}, \binits{A.}},
\beditor{\bsnm{Srikumar}, \binits{V.}} (eds.)
\bbtitle{Proceedings of the 62nd Annual Meeting of the Association for Computational Linguistics (Volume 1: Long Papers)},
pp. \bfpage{6106}--\blpage{6131}.
\bpublisher{Association for Computational Linguistics},
\blocation{Bangkok, Thailand}
(\byear{2024}).
\doiurl{10.18653/v1/2024.acl-long.331} .
\burl{https://aclanthology.org/2024.acl-long.331}
\end{bchapter}
\endbibitem

\bibitem[\protect\citeauthoryear{Zhang et~al.}{2024}]{zhang-etal-2024-exploring}
\begin{bchapter}
\bauthor{\bsnm{Zhang}, \binits{J.}},
\bauthor{\bsnm{Xu}, \binits{X.}},
\bauthor{\bsnm{Zhang}, \binits{N.}},
\bauthor{\bsnm{Liu}, \binits{R.}},
\bauthor{\bsnm{Hooi}, \binits{B.}},
\bauthor{\bsnm{Deng}, \binits{S.}}:
\bctitle{Exploring collaboration mechanisms for {LLM} agents: A social psychology view}.
In: \beditor{\bsnm{Ku}, \binits{L.-W.}},
\beditor{\bsnm{Martins}, \binits{A.}},
\beditor{\bsnm{Srikumar}, \binits{V.}} (eds.)
\bbtitle{Proceedings of the 62nd Annual Meeting of the Association for Computational Linguistics (Volume 1: Long Papers)},
pp. \bfpage{14544}--\blpage{14607}.
\bpublisher{Association for Computational Linguistics},
\blocation{Bangkok, Thailand}
(\byear{2024}).
\doiurl{10.18653/v1/2024.acl-long.782} .
\burl{https://aclanthology.org/2024.acl-long.782}
\end{bchapter}
\endbibitem

\bibitem[\protect\citeauthoryear{AI@Meta}{2024}]{llama31modelcard}
\begin{botherref}
\oauthor{\bsnm{AI@Meta}}:
Llama 3.1 Model Card
(2024).
\url{https://github.com/meta-llama/llama-models/blob/main/models/llama3\_1\\/MODEL\_CARD.md}
\end{botherref}
\endbibitem

\bibitem[\protect\citeauthoryear{Jiang et~al.}{2023}]{jiang2023mistral7b}
\begin{botherref}
\oauthor{\bsnm{Jiang}, \binits{A.Q.}},
\oauthor{\bsnm{Sablayrolles}, \binits{A.}},
\oauthor{\bsnm{Mensch}, \binits{A.}},
\oauthor{\bsnm{Bamford}, \binits{C.}},
\oauthor{\bsnm{Chaplot}, \binits{D.S.}},
\oauthor{\bsnm{Casas}, \binits{D.}},
\oauthor{\bsnm{Bressand}, \binits{F.}},
\oauthor{\bsnm{Lengyel}, \binits{G.}},
\oauthor{\bsnm{Lample}, \binits{G.}},
\oauthor{\bsnm{Saulnier}, \binits{L.}},
\oauthor{\bsnm{Lavaud}, \binits{L.R.}},
\oauthor{\bsnm{Lachaux}, \binits{M.-A.}},
\oauthor{\bsnm{Stock}, \binits{P.}},
\oauthor{\bsnm{Scao}, \binits{T.L.}},
\oauthor{\bsnm{Lavril}, \binits{T.}},
\oauthor{\bsnm{Wang}, \binits{T.}},
\oauthor{\bsnm{Lacroix}, \binits{T.}},
\oauthor{\bsnm{Sayed}, \binits{W.E.}}:
Mistral 7B
(2023).
\url{https://arxiv.org/abs/2310.06825}
\end{botherref}
\endbibitem

\bibitem[\protect\citeauthoryear{Team}{2024a}]{gemmateam2024gemma2improvingopen}
\begin{botherref}
\oauthor{\bsnm{Team}, \binits{G.}}:
Gemma 2: Improving Open Language Models at a Practical Size
(2024).
\url{https://arxiv.org/abs/2403.05530}
\end{botherref}
\endbibitem

\bibitem[\protect\citeauthoryear{Team}{2024b}]{qwen2.5}
\begin{botherref}
\oauthor{\bsnm{Team}, \binits{Q.}}:
Qwen2.5: A Party of Foundation Models
(2024).
\url{https://qwenlm.github.io/blog/qwen2.5/}
\end{botherref}
\endbibitem

\bibitem[\protect\citeauthoryear{Tunstall et~al.}{2023}]{tunstall2023zephyrdirectdistillationlm}
\begin{botherref}
\oauthor{\bsnm{Tunstall}, \binits{L.}},
\oauthor{\bsnm{Beeching}, \binits{E.}},
\oauthor{\bsnm{Lambert}, \binits{N.}},
\oauthor{\bsnm{Rajani}, \binits{N.}},
\oauthor{\bsnm{Rasul}, \binits{K.}},
\oauthor{\bsnm{Belkada}, \binits{Y.}},
\oauthor{\bsnm{Huang}, \binits{S.}},
\oauthor{\bsnm{Werra}, \binits{L.}},
\oauthor{\bsnm{Fourrier}, \binits{C.}},
\oauthor{\bsnm{Habib}, \binits{N.}},
\oauthor{\bsnm{Sarrazin}, \binits{N.}},
\oauthor{\bsnm{Sanseviero}, \binits{O.}},
\oauthor{\bsnm{Rush}, \binits{A.M.}},
\oauthor{\bsnm{Wolf}, \binits{T.}}:
Zephyr: Direct Distillation of LM Alignment
(2023).
\url{https://arxiv.org/abs/2310.16944}
\end{botherref}
\endbibitem

\bibitem[\protect\citeauthoryear{Microsoft}{2024}]{abdin2024phi3technicalreporthighly}
\begin{botherref}
\oauthor{\bsnm{Microsoft}}:
Phi-3 Technical Report: A Highly Capable Language Model Locally on Your Phone
(2024).
\url{https://arxiv.org/abs/2404.14219}
\end{botherref}
\endbibitem

\bibitem[\protect\citeauthoryear{Team}{2024}]{geminiteam2024gemini15unlockingmultimodal}
\begin{botherref}
\oauthor{\bsnm{Team}, \binits{G.}}:
Gemini 1.5: Unlocking multimodal understanding across millions of tokens of context
(2024).
\url{https://arxiv.org/abs/2403.05530}
\end{botherref}
\endbibitem

\bibitem[\protect\citeauthoryear{Anthropic}{2024}]{anthropic2024claude35}
\begin{botherref}
\oauthor{\bsnm{Anthropic}}:
Claude 3.5 sonnet model card addendum.
Technical report,
Anthropic
(September 2024).
Technical Report.
\url{https://www.anthropic.com/modelcards/claude-35-sonnet-addendum}
\end{botherref}
\endbibitem

\bibitem[\protect\citeauthoryear{OpenAI}{2024}]{gpt40modelcard}
\begin{botherref}
\oauthor{\bsnm{OpenAI}}:
GPT-4o Model Card
(2024).
\url{https://openai.com/index/hello-gpt-4o/}
\end{botherref}
\endbibitem

\bibitem[\protect\citeauthoryear{Devlin et~al.}{2019}]{devlin-etal-2019-bert}
\begin{bchapter}
\bauthor{\bsnm{Devlin}, \binits{J.}},
\bauthor{\bsnm{Chang}, \binits{M.-W.}},
\bauthor{\bsnm{Lee}, \binits{K.}},
\bauthor{\bsnm{Toutanova}, \binits{K.}}:
\bctitle{{BERT}: Pre-training of deep bidirectional transformers for language understanding}.
In: \beditor{\bsnm{Burstein}, \binits{J.}},
\beditor{\bsnm{Doran}, \binits{C.}},
\beditor{\bsnm{Solorio}, \binits{T.}} (eds.)
\bbtitle{Proceedings of the 2019 Conference of the North {A}merican Chapter of the Association for Computational Linguistics: Human Language Technologies, Volume 1 (Long and Short Papers)},
pp. \bfpage{4171}--\blpage{4186}.
\bpublisher{Association for Computational Linguistics},
\blocation{Minneapolis, Minnesota}
(\byear{2019}).
\doiurl{10.18653/v1/N19-1423} .
\burl{https://aclanthology.org/N19-1423}
\end{bchapter}
\endbibitem

\bibitem[\protect\citeauthoryear{Conneau et~al.}{2020}]{conneau-etal-2020-unsupervised}
\begin{bchapter}
\bauthor{\bsnm{Conneau}, \binits{A.}},
\bauthor{\bsnm{Khandelwal}, \binits{K.}},
\bauthor{\bsnm{Goyal}, \binits{N.}},
\bauthor{\bsnm{Chaudhary}, \binits{V.}},
\bauthor{\bsnm{Wenzek}, \binits{G.}},
\bauthor{\bsnm{Guzm{\'a}n}, \binits{F.}},
\bauthor{\bsnm{Grave}, \binits{E.}},
\bauthor{\bsnm{Ott}, \binits{M.}},
\bauthor{\bsnm{Zettlemoyer}, \binits{L.}},
\bauthor{\bsnm{Stoyanov}, \binits{V.}}:
\bctitle{Unsupervised cross-lingual representation learning at scale}.
In: \beditor{\bsnm{Jurafsky}, \binits{D.}},
\beditor{\bsnm{Chai}, \binits{J.}},
\beditor{\bsnm{Schluter}, \binits{N.}},
\beditor{\bsnm{Tetreault}, \binits{J.}} (eds.)
\bbtitle{Proceedings of the 58th Annual Meeting of the Association for Computational Linguistics},
pp. \bfpage{8440}--\blpage{8451}.
\bpublisher{Association for Computational Linguistics},
\blocation{Online}
(\byear{2020}).
\doiurl{10.18653/v1/2020.acl-main.747} .
\burl{https://aclanthology.org/2020.acl-main.747}
\end{bchapter}
\endbibitem

\bibitem[\protect\citeauthoryear{Wang et~al.}{2024}]{WANG2024106315}
\begin{barticle}
\bauthor{\bsnm{Wang}, \binits{B.}},
\bauthor{\bsnm{Li}, \binits{X.}},
\bauthor{\bsnm{Yang}, \binits{Z.}},
\bauthor{\bsnm{Guan}, \binits{Y.}},
\bauthor{\bsnm{Li}, \binits{J.}},
\bauthor{\bsnm{Wang}, \binits{S.}}:
\batitle{Unsupervised sentence representation learning with frequency-induced adversarial tuning and incomplete sentence filtering}.
\bjtitle{Neural Networks}
\bvolume{175},
\bfpage{106315}
(\byear{2024})
\doiurl{10.1016/j.neunet.2024.106315}
\end{barticle}
\endbibitem

\bibitem[\protect\citeauthoryear{Bai et~al.}{2024}]{bai2024mt}
\begin{bchapter}
\bauthor{\bsnm{Bai}, \binits{G.}},
\bauthor{\bsnm{Liu}, \binits{J.}},
\bauthor{\bsnm{Bu}, \binits{X.}},
\bauthor{\bsnm{He}, \binits{Y.}},
\bauthor{\bsnm{Liu}, \binits{J.}},
\bauthor{\bsnm{Zhou}, \binits{Z.}},
\bauthor{\bsnm{Lin}, \binits{Z.}},
\bauthor{\bsnm{Su}, \binits{W.}},
\bauthor{\bsnm{Ge}, \binits{T.}},
\bauthor{\bsnm{Zheng}, \binits{B.}},
\bauthor{\bsnm{Ouyang}, \binits{W.}}:
\bctitle{{MT}-bench-101: A fine-grained benchmark for evaluating large language models in multi-turn dialogues}.
In: \beditor{\bsnm{Ku}, \binits{L.-W.}},
\beditor{\bsnm{Martins}, \binits{A.}},
\beditor{\bsnm{Srikumar}, \binits{V.}} (eds.)
\bbtitle{Proceedings of the 62nd Annual Meeting of the Association for Computational Linguistics (Volume 1: Long Papers)},
pp. \bfpage{7421}--\blpage{7454}.
\bpublisher{Association for Computational Linguistics},
\blocation{Bangkok, Thailand}
(\byear{2024}).
\doiurl{10.18653/v1/2024.acl-long.401} .
\burl{https://aclanthology.org/2024.acl-long.401}
\end{bchapter}
\endbibitem

\bibitem[\protect\citeauthoryear{Hu et~al.}{2022}]{hu2022lora}
\begin{bchapter}
\bauthor{\bsnm{Hu}, \binits{E.J.}},
\bauthor{\bsnm{shen}},
\bauthor{\bsnm{Wallis}, \binits{P.}},
\bauthor{\bsnm{Allen-Zhu}, \binits{Z.}},
\bauthor{\bsnm{Li}, \binits{Y.}},
\bauthor{\bsnm{Wang}, \binits{S.}},
\bauthor{\bsnm{Wang}, \binits{L.}},
\bauthor{\bsnm{Chen}, \binits{W.}}:
\bctitle{Lo{RA}: Low-rank adaptation of large language models}.
In: \bbtitle{International Conference on Learning Representations}
(\byear{2022}).
\burl{https://openreview.net/forum?id=nZeVKeeFYf9}
\end{bchapter}
\endbibitem

\bibitem[\protect\citeauthoryear{Reimers and Gurevych}{2019}]{reimers-gurevych-2019-sentence}
\begin{bchapter}
\bauthor{\bsnm{Reimers}, \binits{N.}},
\bauthor{\bsnm{Gurevych}, \binits{I.}}:
\bctitle{Sentence-{BERT}: Sentence embeddings using {S}iamese {BERT}-networks}.
In: \beditor{\bsnm{Inui}, \binits{K.}},
\beditor{\bsnm{Jiang}, \binits{J.}},
\beditor{\bsnm{Ng}, \binits{V.}},
\beditor{\bsnm{Wan}, \binits{X.}} (eds.)
\bbtitle{Proceedings of the 2019 Conference on Empirical Methods in Natural Language Processing and the 9th International Joint Conference on Natural Language Processing (EMNLP-IJCNLP)},
pp. \bfpage{3982}--\blpage{3992}.
\bpublisher{Association for Computational Linguistics},
\blocation{Hong Kong, China}
(\byear{2019}).
\doiurl{10.18653/v1/D19-1410} .
\burl{https://aclanthology.org/D19-1410}
\end{bchapter}
\endbibitem

\end{thebibliography}

\end{document}